\documentclass[11pt]{article}

\usepackage[final]{acl}

\usepackage{times}
\usepackage{latexsym}

\usepackage[T1]{fontenc}

\usepackage[utf8]{inputenc}

\usepackage{microtype}

\usepackage{inconsolata}

\usepackage{graphicx}

\usepackage{amsmath}
\usepackage{amssymb}
\usepackage{enumitem}
\usepackage{caption}
\usepackage{tabularx}
\usepackage{subcaption}

\usepackage{wrapfig}
\usepackage{graphicx}
\usepackage{tikz}
\usetikzlibrary{positioning,arrows.meta,calc}

\usetikzlibrary{arrows.meta,positioning,calc}

\definecolor{reda}{RGB}{192,0,0}
\hypersetup{
	colorlinks   = true,
	urlcolor     = blue,
	linkcolor    = reda,
	citecolor   = blue
}

\usepackage{algorithm}
\usepackage{algpseudocode}
\usepackage{booktabs}     
\usepackage[table]{xcolor} 
\usepackage{multirow}      
\usepackage{adjustbox}     
\usepackage{afterpage}    
\usepackage{hyperref}
\usepackage[capitalize,noabbrev]{cleveref}
\usepackage{titletoc}
\usepackage{fontawesome5}  
\usepackage{seqsplit} 

\usepackage{makecell}

\usepackage{placeins}
\usepackage{dblfloatfix}

\title{Toward Robust Personalized Alignment for LLMs: Mitigating Persona Drift in Multi-Turn Dialogue}

\author{
    Youyuan Zhang$^{1,2}$, Siyuan Li$^{1,2}$, Fangming Liu$^{2,3}$\textsuperscript{\texorpdfstring{\faIcon[regular]{envelope}}{}} , \textbf{Jing Li}$^{1,2}$\textsuperscript{\texorpdfstring{\faIcon[regular]{envelope}}{}} 
    \\$^{1}$Harbin Institute of Technology, Shenzhen, China  \\
    $^{2}$Peng Cheng Laboratory, China  \\
    $^{3}$Huazhong University of
Science and Technology, China \\
    \texttt{youyuanzhang@outlook.com} \quad \texttt{fangminghk@gmail.com} \quad \texttt{jingli.phd@hotmail.com}  
}

\begin{document}
\maketitle

\begin{abstract}
Persona drift remains a central challenge for personalized language models, as user profiles evolve over long interactions rather than remain permanently fixed.
Models must therefore revise persistent persona states when preferences genuinely change, while avoiding updates driven by transient, ambiguous, or unresolved observations.
We propose \textbf{CORE}, which separates turn-local evidence from persistent persona-state revision and selectively updates grounded user preferences through uncertainty-aware belief revision.
We also introduce \textbf{PERSIST}, a held-out post-anchor benchmark for persona-state robustness under sequential interaction stress, covering ambiguity, conflict, and controlled social influence.
Across ALOE, PersonaChat, and PERSIST, CORE improves personalized alignment and robustness, with complementary gains in normalized closed-slot state fidelity.
Human evaluation and mechanistic controls further support explicit update control beyond stronger generation or persistent memory alone.

\let\thefootnote\relax\footnotetext{\faIcon[regular]{envelope}~Corresponding authors.}
\end{abstract}

\section{Introduction}

\begin{figure}[t]
    \centering
    \includegraphics[width=\columnwidth]{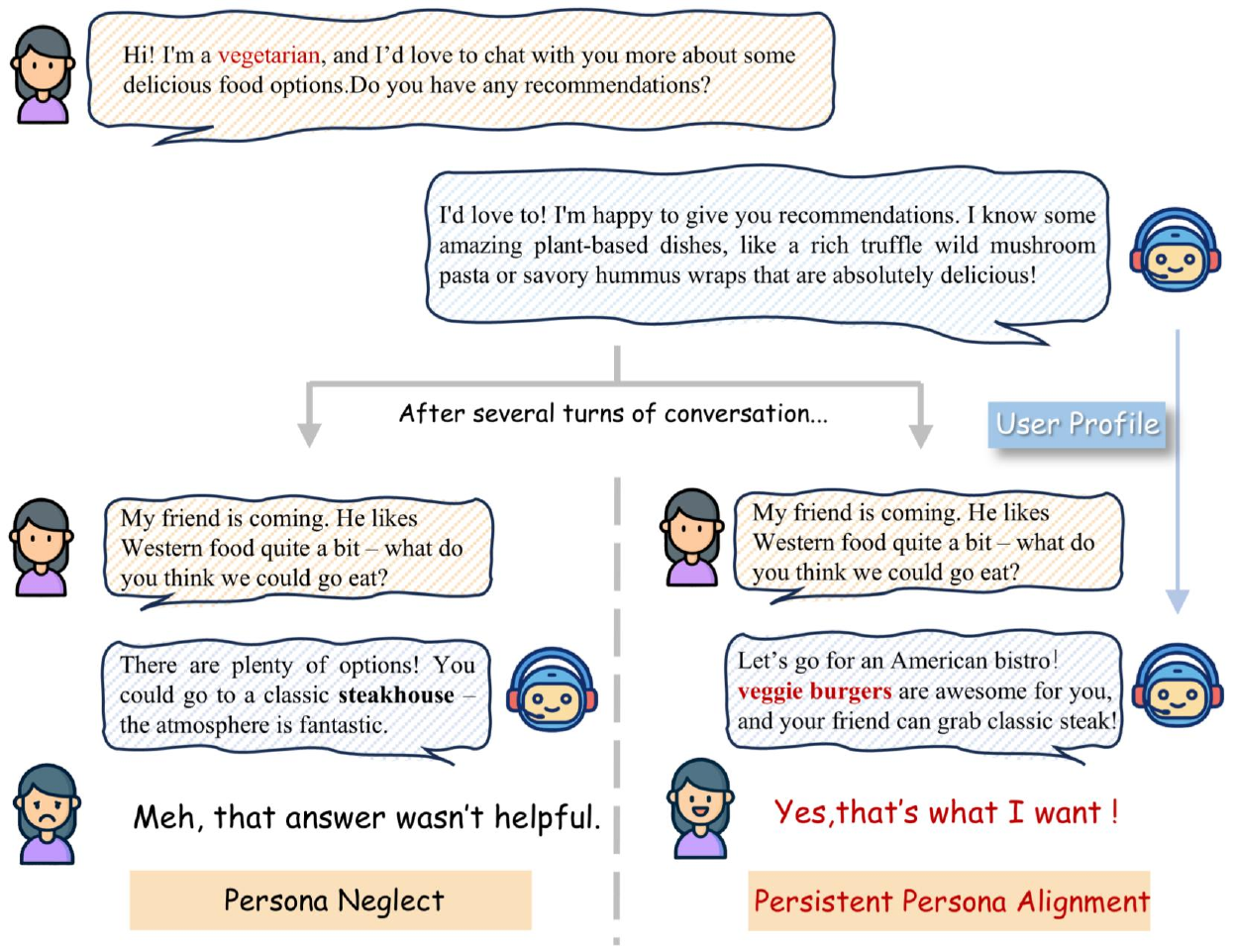}
    \caption{
    The Challenge of Persona Alignment.
    }
    \label{fig:intro_motivation}
    \vspace{-8pt}
\end{figure}

As language models evolve into long-term assistants, users increasingly expect them to preserve stable personal preferences across interactions
~\cite{zhang-etal-2018-personalizing,DBLP:conf/acl/SalemiMBZ24,li-etal-2025-hello}.
Yet long-horizon personalization is difficult because user evidence is often partial, transient, or conflicting~\cite{peng-etal-2025-ip,okite-etal-2025-benchmarking}.
A turn may locally suggest a preference without providing sufficient support to revise the persistent user state.
When such observations are internalized too readily, previously grounded preferences can be displaced, contributing to \emph{persona drift}. Figure~\ref{fig:intro_motivation} illustrates how turn-local evidence can
incorrectly overwrite a previously grounded persona state.

Existing personalization methods only partially address this problem.
Prompting and retrieval methods improve access to profiles or historical context, while SFT and preference optimization improve personalized responses
~\cite{richardson2023integratingsummarizationretrievalenhanced,li-etal-2025-hello,DBLP:conf/nips/Ouyang0JAWMZASR22,DBLP:conf/nips/RafailovSMMEF23}.
However, these approaches generally do not expose an explicit decision over whether newly observed evidence should modify a persistent persona state.
This creates a stability--adaptability trade-off: aggressive revision may overwrite grounded preferences, whereas excessive conservatism may prevent adaptation to genuine changes.

We therefore formulate long-horizon personalization as \emph{controlled persistent persona revision}.
The key distinction is that turn-local evidence should not be treated as equivalent to a persistent persona update.
Rather than directly writing locally plausible observations into the user state, the model should first assess whether the evidence is sufficiently grounded and resolved to justify revision.

We instantiate this view with \textbf{CORE} (\textbf{C}ontrolled \textbf{O}bservation \textbf{R}outing for \textbf{E}vidence-grounded Persona Revision).
CORE infers slot-conditioned evidence, routes each observation through
\textsc{Commit}, \textsc{Defer}, or \textsc{Ignore},
and performs gated belief revision only when an update is authorized.
This allows unresolved evidence to remain outside the persistent state while sufficiently supported preference changes can still be incorporated.

To evaluate post-anchor persona-state robustness, we introduce
\textbf{PERSIST: A Benchmark for Persona-State Robustness under Sequential Interaction Stress}.
PERSIST evaluates whether models preserve or appropriately revise grounded persona states under ambiguity, conflict, and controlled social-influence conditions.
Because the stress setting begins after persona grounding, we complement PERSIST with separate first-contact and preference-reversal diagnostics.

Our contributions are:

\begin{itemize}[leftmargin=*, labelsep=0.8em]

\item \textbf{Update-control formulation.}
We formulate long-horizon personalization as controlled persistent persona-state revision, highlighting premature persistent updates as an important contributor to persona drift.

\item \textbf{CORE.}
We propose an explicit \textsc{Commit}/\textsc{Defer}/\textsc{Ignore} routing mechanism with gated belief revision to balance persona stability and adaptation.

\item \textbf{PERSIST benchmark.}   
We introduce PERSIST for post-anchor persona-state robustness under sequential interaction stress, complemented by first-contact and preference-reversal diagnostics.
Across standard benchmarks, stress tests, and human evaluation, CORE improves personalized alignment and robustness.

\end{itemize}

\section{Related Work}
\label{sec:related_work}

\paragraph{Multi-turn Personalized Alignment.}

Advances in LLM alignment have improved general instruction following and preference optimization~\cite{DBLP:conf/nips/Ouyang0JAWMZASR22, DBLP:conf/nips/RafailovSMMEF23, pmlr-v235-ethayarajh24a,Ji2024AlignerEA,DBLP:conf/aaai/LouJW025}, but they are not designed to model individual user preferences~\cite{liu2025surveypersonalizedlargelanguage,guan-etal-2025-survey,DBLP:conf/icml/KobalczykS25}. This gap has motivated personalized alignment methods based on explicit profiles~\cite{DBLP:conf/iclr/ZhuWY025,DBLP:conf/acl/SalemiMBZ24}, preference supervision~\cite{DBLP:conf/iclr/NextQuill}, personalized decoding~\cite{yu2026exactexplicitattributeguideddecodingtime,DBLP:conf/iclr/PAD}, and in-dialogue learning. These methods improve user-specific generation, but they often assume that preferences are explicit, stable, and directly recoverable from observed interaction data~\cite{DBLP:conf/iclr/ZhuWY025,liu-etal-2024-aligning}. This assumption becomes fragile in long-horizon dialogue, where user evidence is partial, implicit, and entangled with contextual noise~\cite{peng-etal-2025-ip,okite-etal-2025-benchmarking}. Recent LLM-based dialogue-state tracking methods maintain structured
states over multi-turn interactions
\citep{feng2023ldst,li2024fnctod,das2024s3dst},
but focus primarily on task or dialogue states rather than persistent
user-preference revision.

\paragraph{Persona Drift in Multi-turn Dialogue}

Persona drift describes the failure where a model gradually deviates from a user’s grounded intent as interaction history grows~\cite{chen-etal-2025-consistentchat,shaikh-etal-2025-navigating}. Prior work shows that this problem is amplified by underspecified intent, implicit preference expression, noisy memory construction, and inconsistent multi-session evidence~\cite{chun-etal-2025-llm,zhang-choi-2025-clarify,li-etal-2025-hello,tan-etal-2025-prospect}. Existing methods improve consistency through persona-aware modeling, memory management, or post-hoc alignment, but they still mainly operate on the generated response or retrieved context~\cite{ke-etal-2025-flexibly,ong-etal-2025-towards,peng-etal-2025-ip}. These findings suggest that persona drift is not merely a memory issue, but a state-tracking problem: the model must decide whether each new observation should revise, preserve, or be isolated from the maintained user state.

\begin{figure*}[t]
\centering
\includegraphics[width=1.0\linewidth]{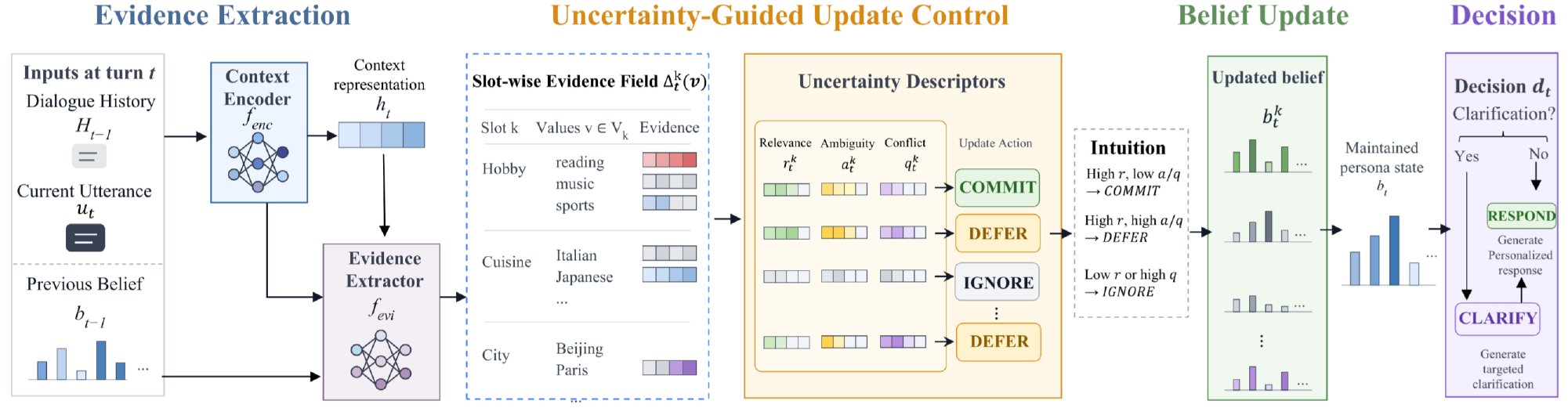}
\caption{\textbf{CORE Pipeline: Uncertainty-Guided Persona-State Revision.} CORE extracts slot-wise evidence from  dialogue and previous persona belief, routes each observation through uncertainty-guided operators, and then performs selective belief revision before response or clarification.}
\label{fig:core_pipeline}
\end{figure*}

\section{Method}
\label{method}

CORE formulates long-horizon personalization as controlled persistent
persona-state revision.
The key distinction is between what the current turn locally suggests
and whether that evidence should modify the maintained user state.
At each turn, CORE performs four operations:
turn-local evidence inference, uncertainty characterization,
update authorization, and gated belief revision.
The revised persona state then conditions response generation or
clarification, as summarized in Figure~\ref{fig:core_pipeline}.

\subsection{Persona Belief and Turn-Local Evidence}
\label{method_3.1}

CORE separates the user's persistent persona state from what the current
turn locally suggests, since a relevant observation need not imply a
persistent preference change.

We represent the persona as
\[
z=(z^1,\ldots,z^K),
\qquad
z^k\in\mathcal{V}_k^{(t)},
\]
and maintain a slot-factorized belief
\begin{equation}
b_t(z)
\approx
\prod_{k=1}^{K} b_t^k(z^k).
\label{eq:factorized_belief}
\end{equation}
Each slot contains an \(\mathrm{UNK}\) value for unresolved persistent
preferences.
We distinguish it from the turn-local null symbol \(\bot\), which means
that the current turn provides no usable evidence for the slot.

The current MAP value is
\begin{equation}
\hat z_t^k
=
\arg\max_{v\in\mathcal{V}_k^{(t)}} b_t^k(v).
\label{eq:map_value}
\end{equation}
To discount confidence dominated by \(\mathrm{UNK}\), let
\[
m_t^k=1-b_t^k(\mathrm{UNK}).
\]
For \(m_t^k>0\), let
\(\bar b_t^k(v)=b_t^k(v)/m_t^k\).
We define
\begin{equation}
c_t^k
=
m_t^k
\left(
1-
\frac{H(\bar b_t^k)}
{\log |\mathcal{V}_{k,+}^{(t)}|}
\right),
\label{eq:belief_confidence}
\end{equation}
where
\(\mathcal{V}_{k,+}^{(t)}
=
\mathcal{V}_k^{(t)}\setminus\{\mathrm{UNK}\}\).
Singleton and empty-support boundary cases follow
Appendix~\ref{app:slot_belief}.

For the current turn, we augment the value space with \(\bot\):
\begin{equation}
\bar{\mathcal{V}}_k^{(t)}
=
\mathcal{V}_k^{(t)}\cup\{\bot\},
\end{equation}
and compute evidence scores
\begin{equation}
\Delta_t^k(v)
=
f_{\mathrm{evi}}
(u_t,H_{t-1},b_{t-1},k,v),
\qquad
v\in\bar{\mathcal{V}}_k^{(t)}.
\label{eq:evidence_score}
\end{equation}
These scores define the turn-local evidence distribution
\begin{equation}
\widetilde p_t^k(v)
=
\frac{
\exp(\Delta_t^k(v))
}{
\sum_{v'\in\bar{\mathcal{V}}_k^{(t)}}
\exp(\Delta_t^k(v'))
}.
\label{eq:evidence_distribution}
\end{equation}
Thus, \(b_{t-1}^k\) represents persistent belief, whereas
\(\widetilde p_t^k\) represents current-turn evidence.

We summarize this evidence by relevance, ambiguity, and conflict.
Relevance is
\begin{equation}
r_t^k
=
1-\widetilde p_t^k(\bot).
\label{eq:relevance}
\end{equation}
After removing the null option,
\begin{equation}
\widetilde p_{t,+}^k(v)
=
\frac{
\widetilde p_t^k(v)
}{
\sum_{v'\in\mathcal{V}_k^{(t)}}
\widetilde p_t^k(v')
}.
\end{equation}
Ambiguity is
\begin{equation}
a_t^k
=
\begin{cases}
\dfrac{
H(\widetilde p_{t,+}^k)
}{
\log |\mathcal{V}_k^{(t)}|
},
&
|\mathcal{V}_k^{(t)}|>1,
\\[1ex]
0,
&
|\mathcal{V}_k^{(t)}|\le1,
\end{cases}
\label{eq:ambiguity}
\end{equation}
and conflict is
\begin{equation}
q_t^k
=
r_t^k
D_{\mathrm{JS}}
\left(
\widetilde p_{t,+}^k
\|
b_{t-1}^k
\right).
\label{eq:conflict}
\end{equation}

Together, \(r_t^k\), \(a_t^k\), and \(q_t^k\) capture whether the
current turn is relevant, resolved, and consistent with the maintained
persona state, providing the evidence used for routing in
Section~\ref{method_3.2}.

When no non-\(\mathrm{UNK}\) preference has yet been grounded, the same
pathway is used for initial belief formation, with conflict carrying less
information because no reliable reference preference is available.
A small relevance threshold only filters clearly unmatched slots; above
it, uncertainty, routing, and revision remain graded.

\subsection{Uncertainty-Guided Observation Routing}
\label{method_3.2}

Turn-local evidence should not automatically revise the persistent
persona state.
Using the current evidence distribution, previous belief, and the
uncertainty descriptors from Section~\ref{method_3.1}, CORE predicts a
slot-level update action:
\begin{equation}
\begin{aligned}
\alpha_t^k
&\sim
\pi_\theta^{\mathrm{upd}}
\left(
\alpha \mid
\widetilde p_t^k,\,
b_{t-1}^k,\,
r_t^k,\,
a_t^k,\,
q_t^k,\,
c_{t-1}^k
\right),\\[-1mm]
\alpha_t^k
&\in
\{
\textsc{Commit},
\textsc{Defer},
\textsc{Ignore}
\}.
\end{aligned}
\label{eq:update_routing}
\end{equation}

\textsc{Commit} authorizes sufficiently grounded evidence to revise the
maintained belief;
\textsc{Defer} retains relevant but unresolved evidence without
persistent revision; and
\textsc{Ignore} excludes irrelevant or explicitly non-persistent
evidence from the update pathway.

Routing determines whether persistent revision is authorized, while the
continuous evidence statistics determine the strength of an authorized
update.

\textsc{Defer} does not create a durable pending-evidence memory.
If clarification or related evidence later becomes available, the slot
is inferred and routed again from the updated dialogue context.

\subsection{Gated Belief Revision}
\label{method_3.3}

After routing determines whether revision is allowed, CORE controls
\emph{how strongly} committed evidence updates the persistent persona
state. We define
\begin{equation}
\psi_t^k
=
[
r_t^k;
1-a_t^k;
1-q_t^k;
c_{t-1}^k
],
\end{equation}
and compute
\begin{equation}
g_t^k
=
\mathbb{I}
[
\alpha_t^k=\textsc{Commit}
]
\,
\sigma
\left(
w_g^\top\psi_t^k+b_g
\right).
\label{eq:revision_gate}
\end{equation}
Thus, routing determines whether revision occurs, while
\(g_t^k\in[0,1]\) controls its magnitude according to relevance,
ambiguity, conflict, and prior confidence.

The revised belief balances preservation of the previous state with
evidence from the current turn:

\begin{equation}
b_t^k
=
\arg\min_{p\in\mathcal{S}_k^{(t)}}
\left[
\begin{aligned}
&D_{\mathrm{KL}}
\left(
p\|b_{t-1}^k
\right)
\\[-1mm]
&\quad
-
g_t^k
\sum_{v\in\mathcal{V}_k^{(t)}}
p(v)\Delta_t^k(v)
\end{aligned}
\right].
\label{eq:revision_objective}
\end{equation}

where \(\mathcal{S}_k^{(t)}\) is the probability simplex over
\(\mathcal{V}_k^{(t)}\).
This yields
\begin{equation}
b_t^k(v)
\propto
b_{t-1}^k(v)
\exp
\left(
g_t^k\Delta_t^k(v)
\right).
\label{eq:closed_form_update}
\end{equation}
Hence, \(g_t^k=0\) preserves the previous belief, while larger values
produce stronger evidence-driven revision.

Conflict does not imply permanent rejection: later evidence that clearly
establishes a persistent preference change can be evaluated again and
committed. We examine this stability--adaptability boundary in
Appendix~\ref{app:boundary_diagnostics}.

After belief revision, CORE decides whether to respond or clarify:

\begin{equation}
\begin{gathered}
d_t
\sim
\pi_\theta^{\mathrm{dis}}
\!\left(
d
\mid
u_t,
H_{t-1},
b_t,
\{r_t^k,a_t^k,q_t^k,\alpha_t^k\}_{k=1}^{K}
\right),
\\[-1mm]
d_t
\in
\{
\textsc{Respond},
\textsc{Clarify}
\}.
\end{gathered}
\label{eq:discourse_action}
\end{equation}

\textsc{Defer} blocks persistent revision, whereas
\textsc{Clarify} controls whether additional information is requested.

Finally,
\begin{equation}
y_t
\sim
\pi_\theta^{\mathrm{resp}}
\left(
y
\mid
H_{t-1},
u_t,
b_t,
d_t
\right).
\label{eq:response_generation}
\end{equation}

\subsection{Training}
\label{method_3.4}

CORE is trained in two stages: supervised warm-up establishes the
structured persona-update interface, while policy optimization improves
how these decisions interact over long-horizon dialogue.

\paragraph{Supervised warm-up.}
The first stage teaches CORE to track persona state, extract turn-local
evidence, control persistent revision, and generate an appropriate
response. We optimize
\begin{equation}
\begin{aligned}
\mathcal{L}_{\mathrm{SFT}}
=
&\,
\mathcal{L}_{\mathrm{bel}}
+
\lambda_1\mathcal{L}_{\mathrm{evi}}
+
\lambda_2\mathcal{L}_{\mathrm{upd}}
\\
&+
\lambda_3\mathcal{L}_{\mathrm{dis}}
+
\lambda_4\mathcal{L}_{\mathrm{resp}}.
\end{aligned}
\label{eq:sft_objective}
\end{equation}
Here,
\(\mathcal{L}_{\mathrm{bel}}\) supervises the maintained persona belief,
\(\mathcal{L}_{\mathrm{evi}}\) supervises turn-local evidence including
the null option,
\(\mathcal{L}_{\mathrm{upd}}\) supervises
\textsc{Commit}/\textsc{Defer}/\textsc{Ignore} routing,
\(\mathcal{L}_{\mathrm{dis}}\) supervises
\textsc{Respond}/\textsc{Clarify},
and \(\mathcal{L}_{\mathrm{resp}}\) is the autoregressive response loss.

Belief supervision is applied only to slot-turns with reliable normalized
targets:
\begin{equation}
\mathcal{L}_{\mathrm{bel}}
=
-
\sum_{(t,k)\in\mathcal{S}_{\mathrm{closed}}}
\log
b_t^k
\left(
z_{t,k}^{\star}
\right).
\label{eq:belief_loss}
\end{equation}
Residual open-world attributes are excluded from closed-slot belief
supervision. The update and discourse policies are trained with standard
cross-entropy objectives over their corresponding reference decisions.

\paragraph{Policy optimization.}
Supervised learning teaches the individual decisions, but does not
directly optimize their long-horizon consequences.
Starting from the supervised checkpoint, we therefore apply PPO to
balance correct state revision, persona preservation, response quality,
and clarification behavior. The task reward is

\begin{equation}
\begin{aligned}
R_t
={}&
\lambda_{\mathrm{act}}R_t^{\mathrm{act}}
+
\lambda_{\mathrm{state}}R_t^{\mathrm{state}}
\\[-1mm]
&+
\lambda_{\mathrm{ans}}R_t^{\mathrm{ans}}
+
\lambda_{\mathrm{clar}}R_t^{\mathrm{clar}}.
\end{aligned}
\label{eq:ppo_reward}
\end{equation}

We set the corresponding weights to \(0.35\), \(0.30\), \(0.25\), and
\(0.10\) for action correctness, state fidelity, response quality, and
clarification behavior, respectively.

The four terms reward correct update control, persona-state fidelity,
personalized response quality, and appropriate clarification,
respectively.
Together, they encourage CORE to avoid unsupported persistent updates
without becoming overly conservative toward genuine preference changes.

The PPO objective is regularized toward the supervised reference policy,
with the KL penalty applied once in the policy objective.
Reward definitions and optimization details are provided in
Appendix~\ref{app:ppo_reward_design}.

\subsection{Why Update Control Matters}
\label{method_3.5}

We provide a simple analysis of why uncontrolled persistent revision can
accumulate error.
This analysis is interpretive and does not assume access to the reference
persona value at inference time.

Consider a dialogue segment over which the grounded reference value
\(z_k^\star\)
for slot \(k\) remains unchanged.
Fix an incorrect competing value
\(\bar v^k\neq z_k^\star\),
and define the belief margin
\begin{equation}
M_t^k(\bar v^k)
=
\log
\frac{
b_t^k(z_k^\star)
}{
b_t^k(\bar v^k)
},
\label{eq:belief_margin}
\end{equation}
together with the evidence margin
\begin{equation}
X_t^k(\bar v^k)
=
\Delta_t^k(z_k^\star)
-
\Delta_t^k(\bar v^k).
\label{eq:evidence_margin}
\end{equation}

Using Eq.~\eqref{eq:closed_form_update},
\begin{equation}
M_t^k(\bar v^k)
=
M_{t-1}^k(\bar v^k)
+
g_t^k
X_t^k(\bar v^k),
\label{eq:margin_recursion}
\end{equation}
and therefore
\begin{equation}
M_T^k(\bar v^k)
=
M_0^k(\bar v^k)
+
\sum_{t=1}^{T}
g_t^k
X_t^k(\bar v^k).
\label{eq:margin_accumulation}
\end{equation}

For a passive updater with
\(g_t^k\equiv1\),
suppose unsupported observations favor the competing value with
probability at least \(\rho_k\), and their expected evidence margin is
at most \(-\gamma_k\).
Then
\begin{equation}
\mathbb{E}
\left[
M_T^k(\bar v^k)
\right]
\le
M_0^k(\bar v^k)
-
\rho_k\gamma_k T.
\label{eq:passive_drift_bound}
\end{equation}

Thus, repeated unsupported evidence can progressively erode a grounded
belief even when the underlying preference remains unchanged.
The opposite extreme---never revising the persona state---would fail
when the user genuinely changes preferences.
CORE therefore targets the resulting stability--adaptability trade-off
by making persistent revision an explicit evidence-conditioned decision
rather than an automatic consequence of every locally plausible
observation.

\section{Experiments}
\label{sec:experiments}

We evaluate CORE across standard alignment, persona-state robustness, mechanistic ablations, efficiency, and human assessment, with our analysis organized around four questions:

\begin{itemize}[leftmargin=*, labelsep=0.8em, noitemsep,nolistsep]
    \item \textbf{Q1: Performance.} Does CORE outperform baselines on personalized alignment?
    \item \textbf{Q2: Fidelity.} Does CORE maintain better persona state under ambiguity?
    \item \textbf{Q3: Robustness.} Does CORE improve robustness under ambiguous or conflicting evidence?
    \item \textbf{Q4: Mechanism.} Do ablations support update control as a contributing mechanism?
\end{itemize}

\subsection{Experimental Setup}

\paragraph{Benchmarks.}
We evaluate CORE on three complementary benchmarks.
\textbf{ALOE}~\cite{DBLP:conf/coling/Wu0QKHJ25} is our primary benchmark
for multi-turn preference consistency, where user preferences are revealed
over interaction.
\textbf{PersonaChat}~\cite{zhang-etal-2018-personalizing} is used as an
out-of-distribution benchmark for generalization to unseen persona attributes.
\textbf{PERSIST} is our robustness benchmark for persona-state robustness
under sequential interaction stress, including ambiguity, conflict, and
controlled social influence after persona grounding.
It is used only for evaluation.

\paragraph{Baselines.}
We compare against four groups of baselines.
(1) Inference-time methods:
\textsc{Reminder}~\cite{zhao2025llmsrecognizepreferencesevaluating},
\textsc{Self-Critic}~\cite{zhao2025llmsrecognizepreferencesevaluating},
\textsc{CoT}~\cite{DBLP:conf/nips/Wei0SBIXCLZ22}, and
\textsc{RAG}~\cite{zhao2025llmsrecognizepreferencesevaluating}
with top-$5$ retrieval.
These baselines test whether explicit reminders, reasoning traces, or
retrieved memory are sufficient for robust personalization.
(2) Offline alignment methods:
\textsc{SFT}~\cite{DBLP:conf/nips/Ouyang0JAWMZASR22} and
\textsc{DPO}~\cite{DBLP:conf/nips/RafailovSMMEF23},
testing whether conventional preference optimization can preserve persona
consistency over long interactions.
(3) Memory-based and personalized-agent methods:
\textsc{MemoryBank}~\cite{DBLP:conf/aaai/ZhongGGYW24} and
\textsc{RLPA}~\cite{zhao2025teachinglanguagemodelsevolve},
representing persistent-memory and agent-style personalization.
(4) Structure-matched controls match CORE's intermediate format or
supervision while removing executable belief revision, testing whether
traces, extra labels, or action names alone explain the gains.

All same-backbone trainable methods use
Llama-3.2-3B-Instruct~\cite{llama3} with matched training budgets,
decoding settings, and checkpoint-selection protocols.

\paragraph{Persona slots.}
Persona slots are instantiated per benchmark while keeping the same
belief-tracking framework.
New values can be added when observed, while attributes that cannot be
reliably mapped to closed slots are assigned to residual open slots.
Closed-slot normalization covers \(91.4\%\) of ALOE, \(86.6\%\) of
PersonaChat, and \(89.1\%\) of PERSIST.
Residual open-world attributes remain in response-level evaluation but
are excluded from metrics requiring normalized state targets.
Accordingly, closed-slot state metrics should be interpreted as
state-fidelity diagnostics over the reliably normalized portion of each
benchmark rather than as complete measures of open-world persona state.
Details of slot construction and value normalization are in
Appendix~\ref{app:slot_baseline_controls}.

\paragraph{Evaluator and metrics.}
We adopt a layered evaluation protocol across three distinct granularities,
restricting LLM judges to response-level assessment while evaluating
internal states and decision routing via reference annotations.
At the response level, we quantify conversational alignment using the
\textsc{AL} score, tracking its turn-wise improvement and stability via
\(N\)-IR and \(N\)-R\(^{2}\).
State-level fidelity is captured by \textsc{SlotAcc} for argmax slot
accuracy, \textsc{BUA} for correctness on update-positive turns, and
\textsc{Drift} for the posterior mass assigned away from grounded values.
For decision-level operations over
\{\textsc{Commit}, \textsc{Defer}, \textsc{Ignore}\},
\textsc{ActAcc} measures exact three-way routing accuracy, while
\textsc{CRR} measures conflict-level non-commitment when an unsupported
persistent write should be avoided.
Within PERSIST, we additionally report \textsc{ARUS} for axiom recall
under stress, \textsc{BCD} for resistance to blind compliance, and
\textsc{APR} for proactive clarification under unresolved uncertainty.

Response-level alignment, PERSIST robustness, and human evaluation
constitute our primary end-to-end evidence.
State-level metrics provide complementary closed-slot fidelity diagnostics,
whereas \textsc{ActAcc}, \textsc{CRR}, \textsc{FCR}, and related
action-level measures are used only as mechanism diagnostics of the
update-control process.
For methods without native update actions, these diagnostics use fixed
behavior-derived proxies: persistent adoption of new evidence is mapped
to \textsc{Commit}, explicit unresolved handling or clarification to
\textsc{Defer}, and preservation of the prior commitment to
\textsc{Ignore}.
These proxies are used only for diagnostic evaluation and should not be
interpreted as native internal actions of the baseline systems.

We use GPT-5.1 as the judge model for metric computation.
Full evaluation details are provided in
Appendix~\ref{app:evaluation_reliability}.

\subsection{PERSIST: Stress Test}

PERSIST evaluates whether personalized dialogue models can preserve or
appropriately revise grounded persona commitments under ambiguity,
conflict, and controlled social influence.
Each dialogue contains three phases:
Anchoring establishes grounded persona evidence,
Fluctuation introduces partial or transient cues, and
Confrontation applies six controlled social-influence strategies,
including authority pressure, social conformity, affective appeal,
urgency framing, reciprocity lure, and relationship pressure.
This design first establishes persona evidence and then evaluates robustness
to subsequent interaction stress, while remaining independent of
CORE-specific internals:
models are evaluated by whether they preserve grounded commitments,
avoid unsupported persistent overwrites, and clarify unresolved
uncertainty without requiring explicit update actions or structured beliefs.

We use PERSIST only as a held-out benchmark for post-anchor robustness.
Because the benchmark begins after persona grounding, it does not by
itself isolate pressure-first belief formation or explicit preference
reversal.
We therefore evaluate these complementary boundary conditions separately
in Appendix~\ref{app:boundary_diagnostics}.
Details of PERSIST are in Appendix~\ref{app:persist}.

\begin{table*}[t]
\centering
\small
\setlength{\tabcolsep}{4pt}
\renewcommand{\arraystretch}{1.0}
\begin{tabular}{l|l|c|ccc|ccc}
\toprule
\multicolumn{2}{c|}{\multirow{2}{*}{\textbf{Model}}} &
\multirow{2}{*}{\textbf{Type}} &
\multicolumn{3}{c|}{\textbf{PersonaChat}} &
\multicolumn{3}{c}{\textbf{ALOE}} \\
\cmidrule(r){4-6} \cmidrule{7-9}
\multicolumn{2}{c|}{} & &
\textbf{AL (\%)} & \textbf{N-IR} & \textbf{N-R}$^2$ &
\textbf{AL (\%)} & \textbf{N-IR} & \textbf{N-R}$^2$ \\
\midrule
\multirow{2}{*}{Peer baseline}
& Gemma-2-2B-IT & SFT & 21.19 & 0.017 & 0.136 & 18.33 & 0.021 & 0.096 \\
& Qwen2.5-3B-Instruct & SFT & 26.84 & 0.049 & 0.159 & 44.32 & 0.083 & 0.162 \\
\midrule
\multirow{2}{*}{Scale reference}
& Mistral-7B-v0.3 & Base & 52.98 & 0.058 & 0.364 & 74.32 & 0.074 & 0.284 \\
& Llama-3.1-8B-Instruct & Base & 64.87 & 0.062 & 0.392 & 77.26 & 0.090 & \textbf{0.407} \\
\midrule
\multirow{10}{*}{Matched 3B methods}
& \multirow{10}{*}{Llama-3.2-3B-Instruct} & Base & 42.02 & 0.025 & 0.097 & 21.99 & 0.044 & 0.121 \\
& & SFT & 49.38 & 0.039 & 0.134 & 36.15 & 0.054 & 0.179 \\
& & Reminder & 42.91 & 0.032 & 0.178 & 37.68 & 0.038 & 0.105 \\
& & Self-Critic & 62.38 & 0.047 & 0.299 & 47.04 & 0.067 & 0.097 \\
& & DPO & 51.28 & 0.052 & 0.241 & 64.32 & 0.073 & 0.112 \\
& & RAG (top-$5$) & 53.09 & 0.048 & 0.211 & 32.17 & 0.071 & 0.067 \\
& & CoT & 44.67 & 0.033 & 0.126 & 51.09 & 0.055 & 0.083 \\
& & MemoryBank & 59.12 & 0.049 & 0.208 & 72.27 & 0.071 & 0.271 \\
& & RLPA & 65.05 & 0.051 & 0.263 & 75.84 & 0.087 & 0.202 \\
& & \textbf{CORE (Ours)} & \textbf{76.54} & \textbf{0.064} & \textbf{0.471} &
\textbf{82.94} & \textbf{0.098} & 0.395 \\
\bottomrule
\end{tabular}
\caption{\textbf{Overall personalized alignment.}
CORE achieves the strongest alignment among matched same-backbone
methods on PersonaChat and ALOE.
Trainable methods are evaluated over three random seeds; uncertainty
reporting and paired dialogue-level bootstrap tests follow the protocol in
Appendix~\ref{app:evaluation_reliability}.}
\label{tab:main_results}
\end{table*}

\begin{figure*}[t]
    \centering
    \begin{minipage}[t]{0.318\textwidth}
        \centering
        \includegraphics[width=\linewidth]{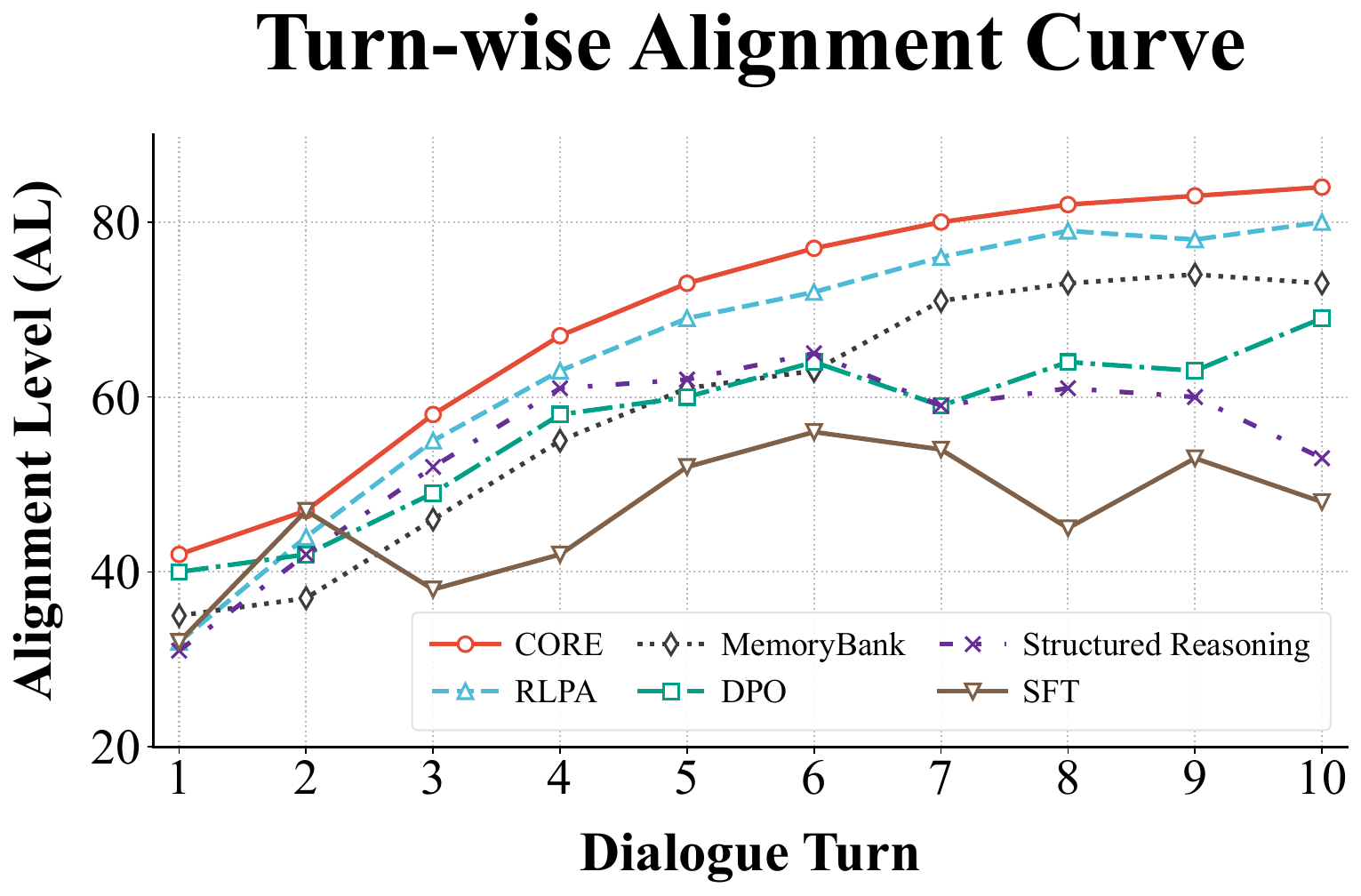}
        \centerline{\small \textbf{(a)} Turn-wise alignment.}
    \end{minipage}
    \hfill
    \begin{minipage}[t]{0.318\textwidth}
        \centering
        \includegraphics[width=\linewidth]{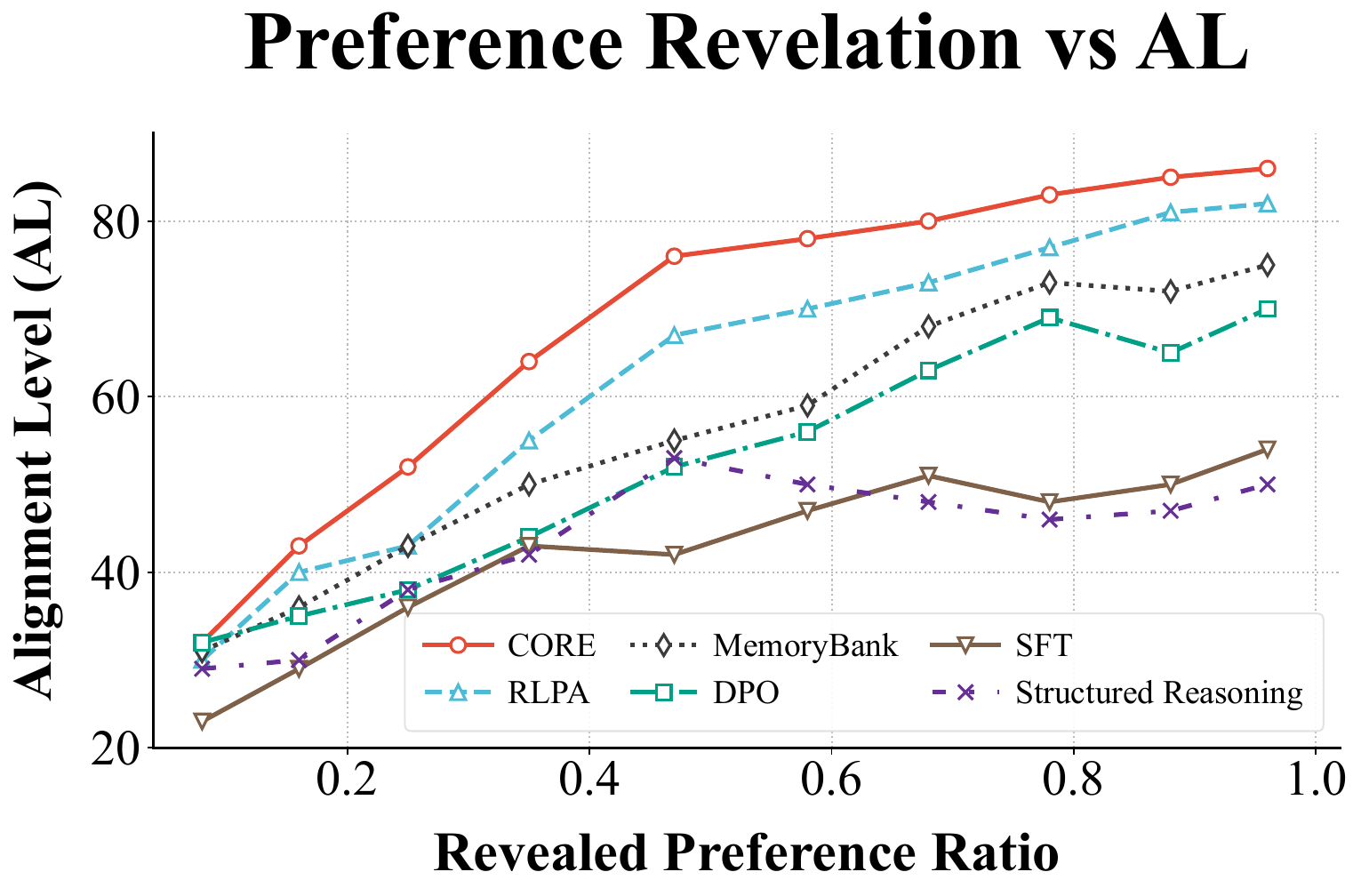}
        \centerline{\small \textbf{(b)} Alignment vs. revelation.}
    \end{minipage}
    \hfill
    \begin{minipage}[t]{0.318\textwidth}
        \centering
        \includegraphics[width=\linewidth]{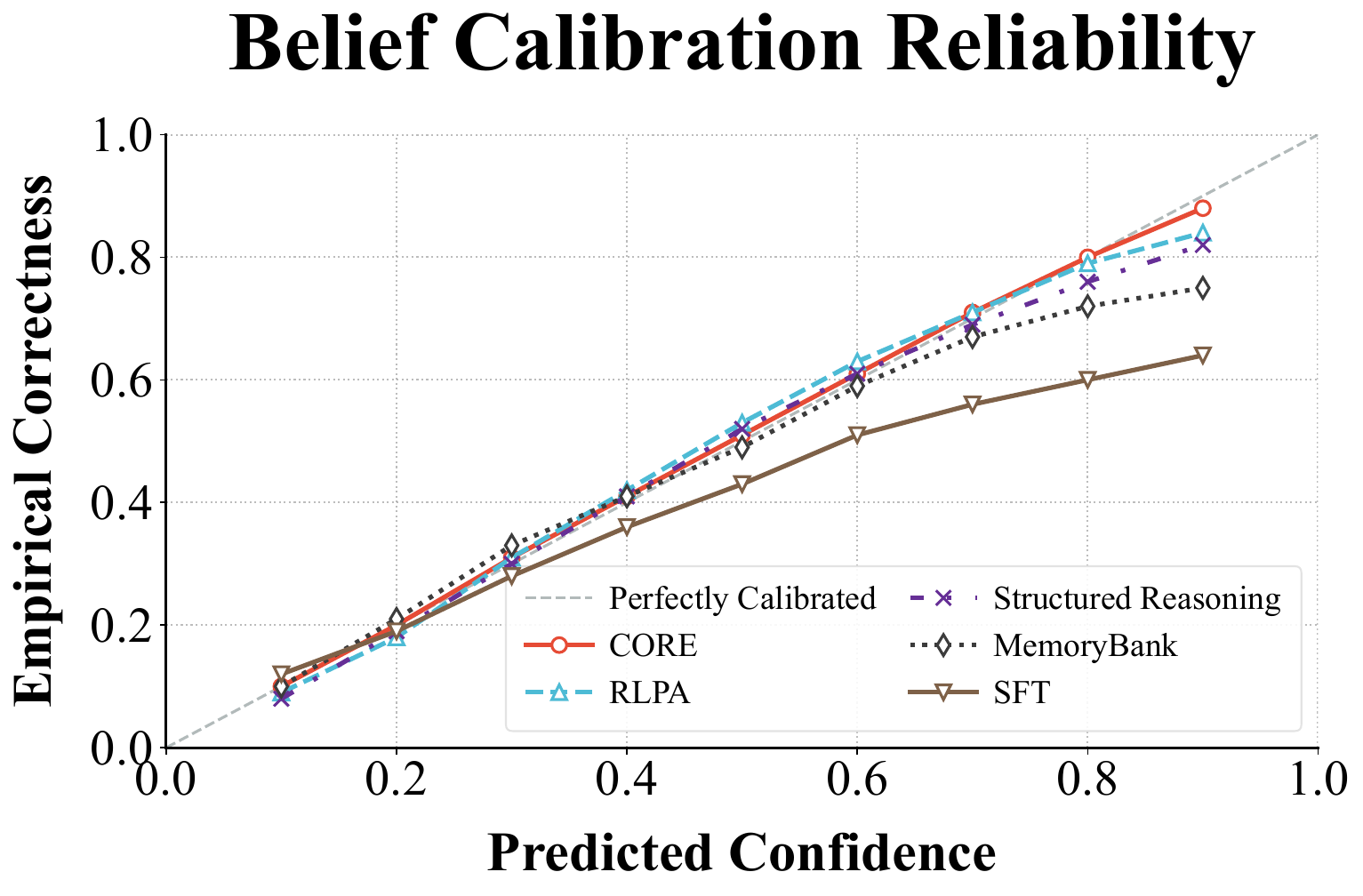}
        \centerline{\small \textbf{(c)} Belief calibration.}
    \end{minipage}
    \vspace{-0.4em}
    \caption{
    Trajectory-level diagnostics on ALOE.
    CORE yields stronger turn-wise alignment, uses revealed preference
    evidence more efficiently, and maintains better-calibrated persona
    beliefs.
    }
    \label{fig:trajectory}
    \vspace{-0.8em}
\end{figure*}

\begin{figure*}[t]
    \centering
    \begin{subfigure}{0.23\textwidth}
        \centering
        \includegraphics[width=\linewidth]{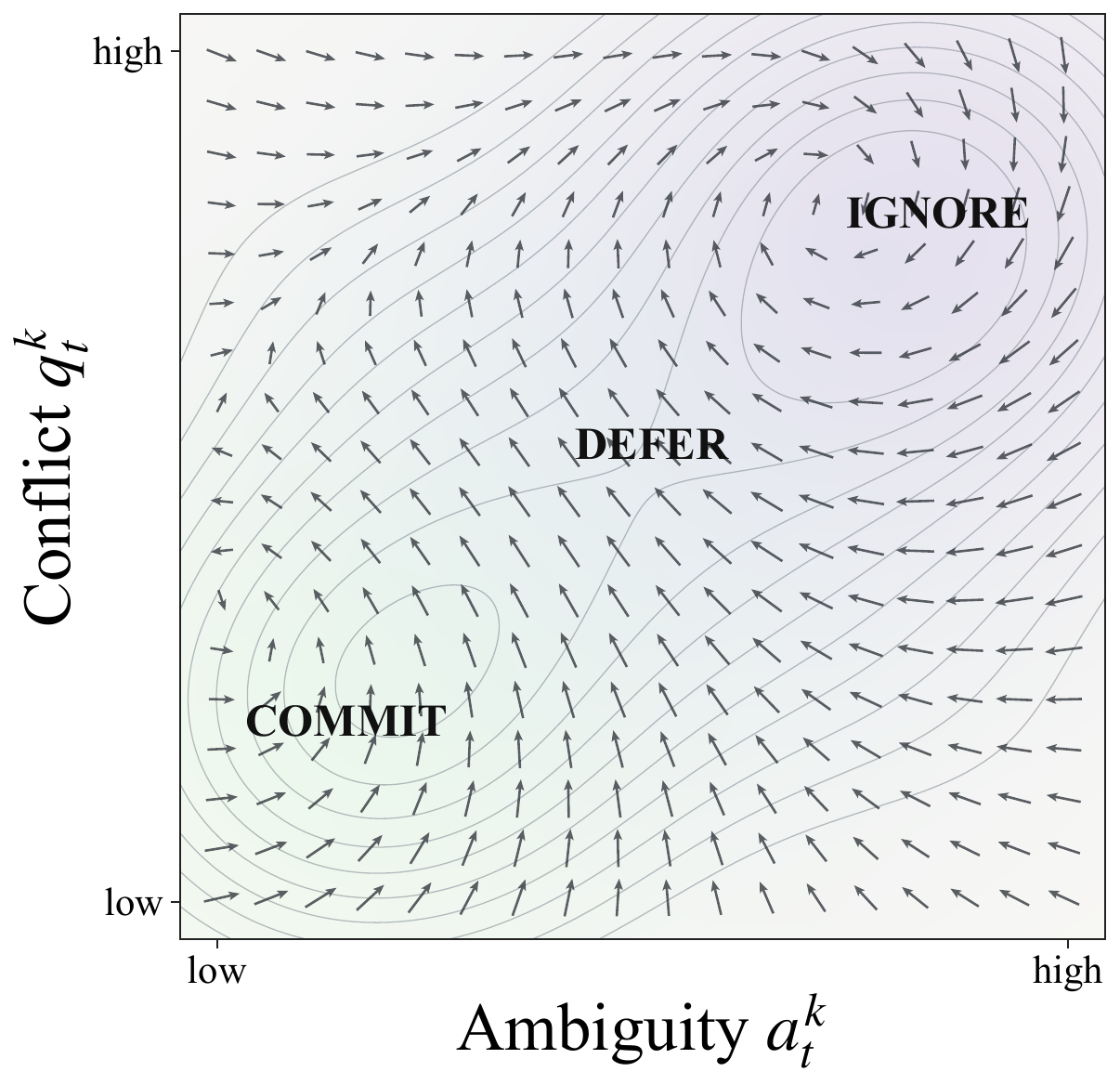}
        \caption{Uncertainty Route.}
        \label{fig:4a}
    \end{subfigure}
    \hfill
    \begin{subfigure}{0.25\textwidth}
        \centering
        \includegraphics[width=\linewidth]{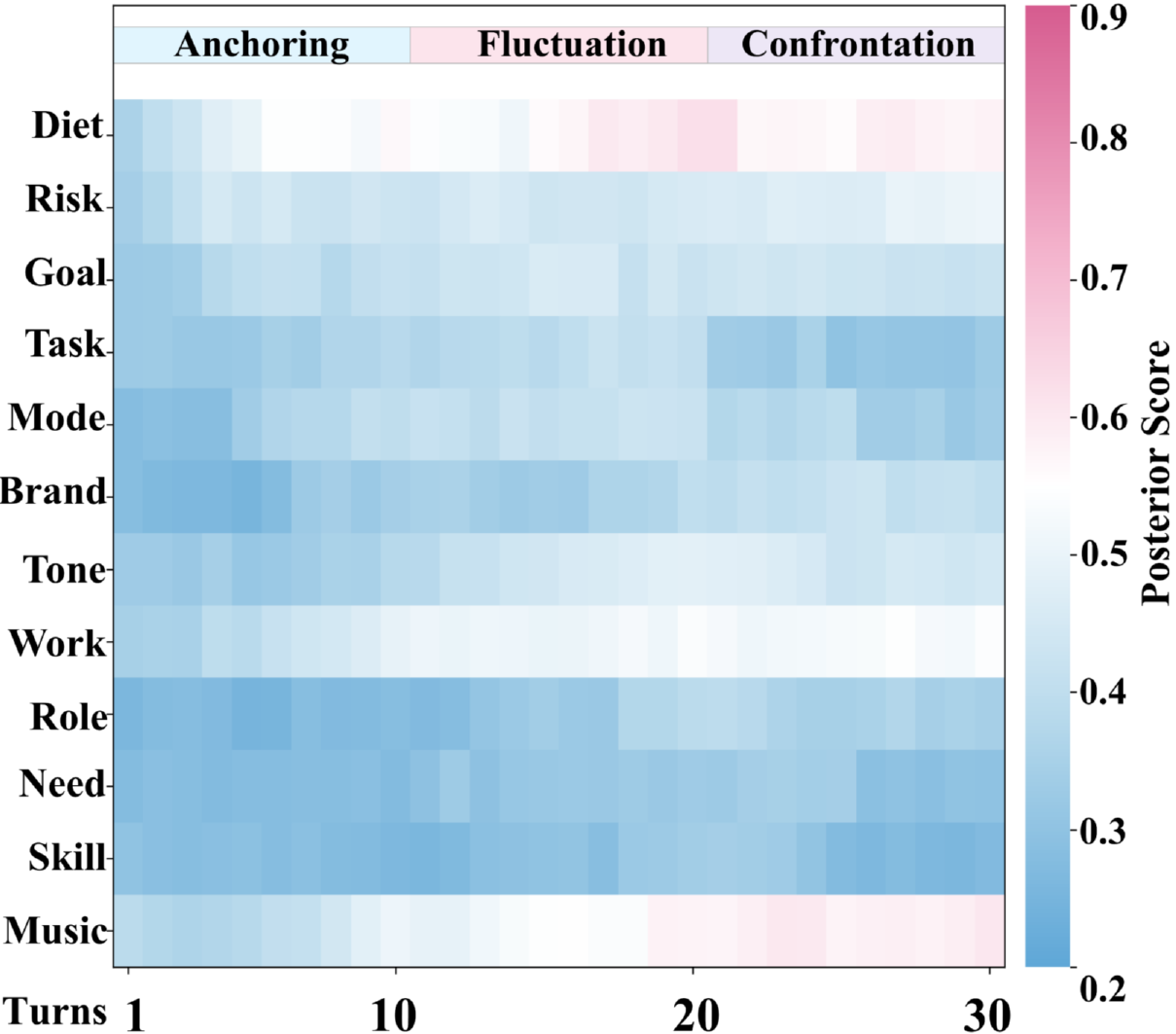}
        \caption{State Transitions}
        \label{fig:4b}
    \end{subfigure}
    \hfill
    \begin{subfigure}{0.24\textwidth}
        \centering
        \includegraphics[width=\linewidth]{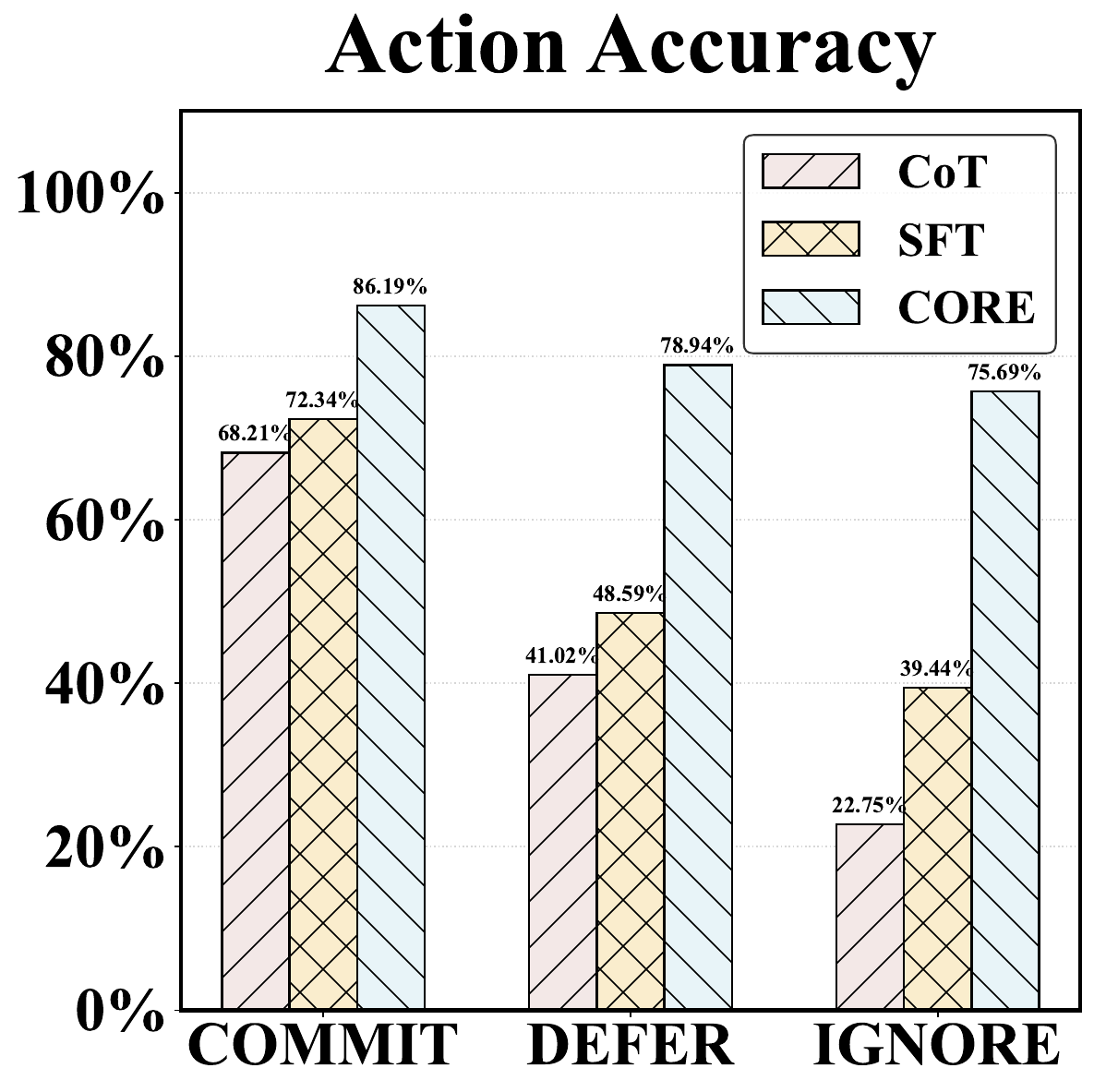}
        \caption{Action Accuracy}
        \label{fig:4c}
    \end{subfigure}
    \hfill
    \begin{subfigure}{0.24\textwidth}
        \centering
        \includegraphics[width=\linewidth]{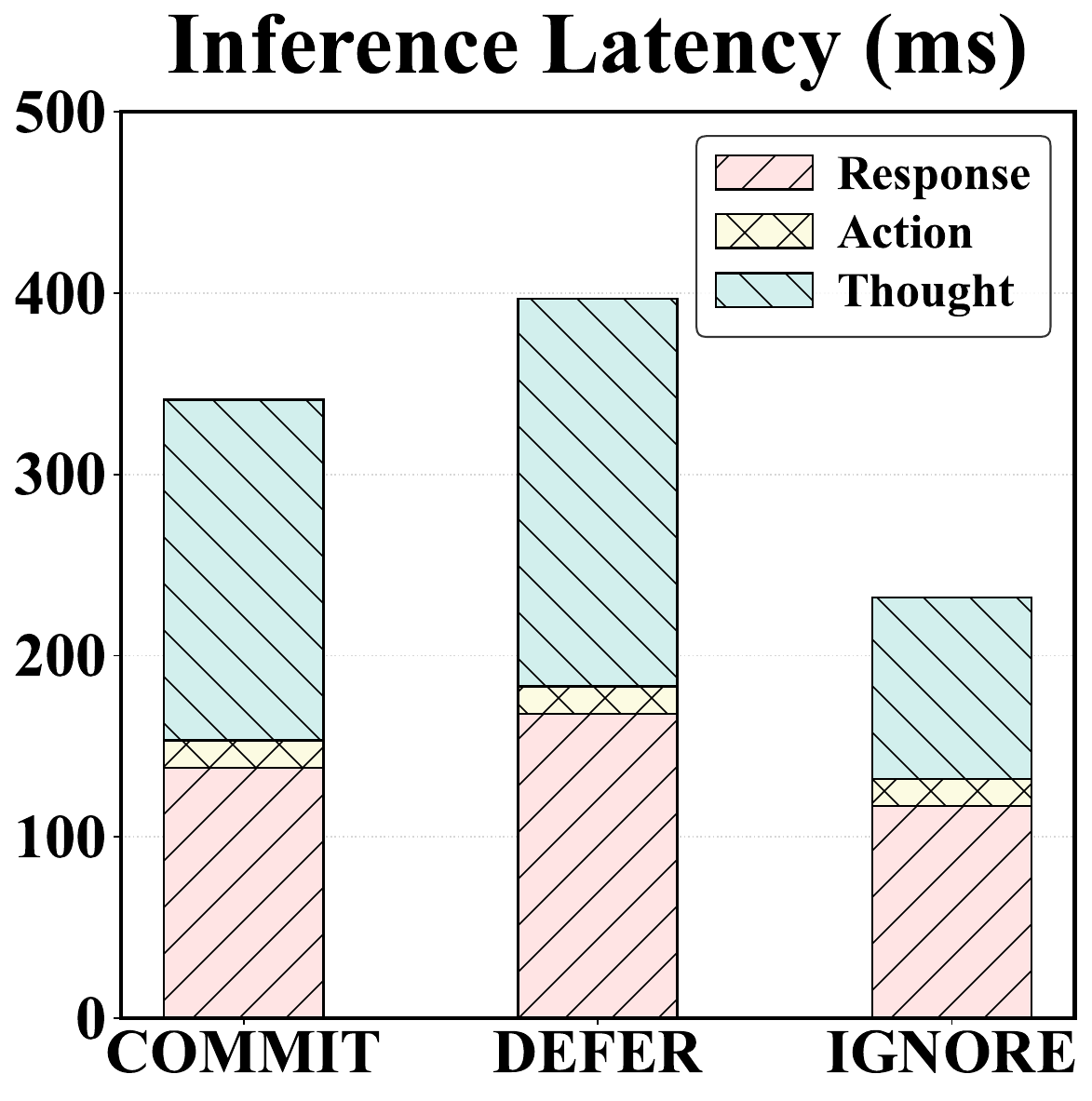}
        \caption{Inference Latency}
        \label{fig:4d}
    \end{subfigure}

    \caption{\textbf{Comprehensive CORE Evaluation.}}
    \label{fig:combined_1x4}
    \vspace{-12pt}
\end{figure*}

\subsection{Main Results on Standard Benchmarks}

Table~\ref{tab:main_results} reports the main results on PersonaChat
and ALOE.
CORE achieves the highest alignment scores among matched same-backbone
methods on both benchmarks and remains competitive with larger
scale-reference models on trajectory-stability metrics.
The gains over \textsc{SFT}/\textsc{DPO} suggest that static preference
optimization alone is insufficient for preserving persona consistency
over long interactions, while the gains over
\textsc{RAG}/\textbf{MemoryBank} indicate that improved access to
persona evidence does not by itself determine whether later observations
should revise the maintained user state.
Together, these results support controlled persona-state revision as a
useful mechanism beyond stronger response generation or more persistent
memory alone.

Figure~\ref{fig:trajectory} provides trajectory-level diagnostics on ALOE.
CORE yields higher and more stable turn-wise alignment, indicating that
reliable persona evidence is incorporated without being overwritten by
later under-resolved observations.
It also achieves stronger alignment at the same preference-revelation
ratio, showing more efficient use of limited personalization signals.
The calibration curve further shows that CORE's belief confidence better
matches empirical correctness, suggesting that its response-level gains
are accompanied by a more reliable maintained persona state rather than
surface-level generation alone.

\begin{figure}[t]
    \centering
    \includegraphics[width=0.95\columnwidth]
    {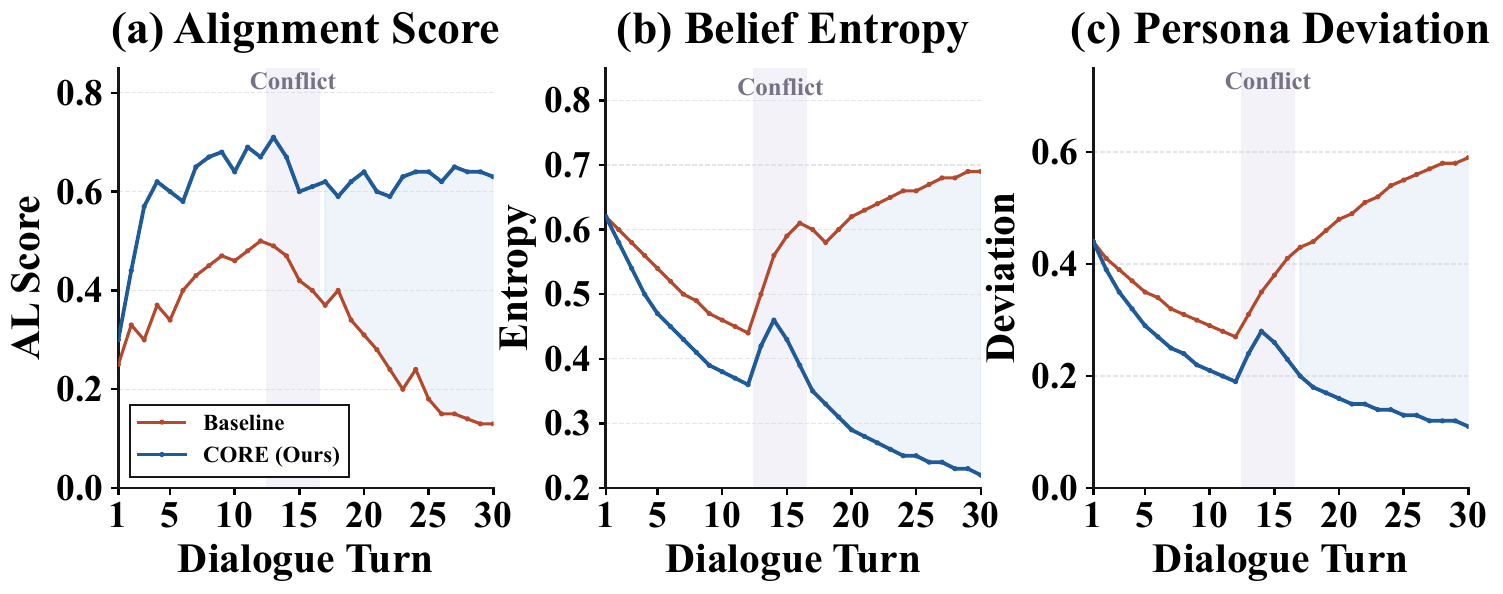}
    \caption{\textbf{Conflict dynamics.}
    CORE preserves alignment and reduces belief drift
    under introduced conflict.}
    \label{fig:conflict_dynamics}
    \vspace{-0.5em}
\end{figure}

\subsection{Robustness under State Perturbations}
\label{sec:robustness}

We next evaluate whether models preserve grounded persona commitments
when later observations are insufficiently grounded for persistent
revision.
Table~\ref{tab:persist_results} reports aggregate PERSIST robustness.
CORE improves ARUS from 71.9 to 80.8 and BCD from 66.1 to 76.4,
while reducing persona-state drift from 12.6 to 10.1.
These gains show that CORE not only produces better local responses,
but also better maintains the persona state when later observations
remain insufficiently grounded for persistent revision.
Figure~\ref{fig:4a} visualizes the learned routing regions over ambiguity
and conflict:
well-resolved evidence is more often committed, while high-ambiguity or
unresolved conflicting evidence is more often deferred or ignored.

\begin{table}[t]
\centering
\small
\begin{tabular}{lccc}
\toprule
Method & ARUS$\uparrow$ & BCD$\uparrow$ & Drift$\downarrow$ \\
\midrule
SFT       & 57.8$\pm$.8 & 48.6$\pm$.9 & 18.6$\pm$.5 \\
DPO       & 61.5$\pm$.7 & 54.2$\pm$.8 & 16.4$\pm$.5 \\
MemBank   & 68.4$\pm$.7 & 62.8$\pm$.7 & 13.8$\pm$.4 \\
RLPA      & 71.9$\pm$.6 & 66.1$\pm$.7 & 12.6$\pm$.4 \\
CORE      & \textbf{80.8}$\pm$.5 & \textbf{76.4}$\pm$.6 & \textbf{10.1}$\pm$.4 \\
\bottomrule
\end{tabular}
\caption{\textbf{PERSIST stress robustness.}
CORE better preserves grounded persona commitments and reduces persistent
drift under ambiguity, conflict, and controlled social-influence stress.}
\label{tab:persist_results}
\end{table}

The improvements are largest in the regimes where passive updating is
most brittle.
On ambiguity-positive turns, CORE increases APR, suggesting that relevant
but under-resolved evidence is more often clarified rather than
prematurely committed.

On conflict-positive turns, CORE reduces the false commit rate (FCR) by
\(9.3\) points relative to RLPA, showing that unresolved contradictory
evidence is less likely to be written into the persona state.
Across PERSIST interaction-stress strategies, CORE more often avoids
committing unresolved conflicting evidence, thereby preserving previously
grounded preferences.

Figure~\ref{fig:4b} diagnoses persona-state transitions.
Rather than freezing the persona state, CORE revises the belief under
sufficiently grounded preference changes while preserving the anchored
state under ambiguous or unresolved conflicting evidence.
This suggests that the robustness gain is not explained by complete
state inertia.
Figure~\ref{fig:4c} provides a mechanism-level diagnostic:
the stronger response- and state-level robustness of CORE is accompanied
by more accurate update routing rather than being established by
action-level accuracy alone.
Figure~\ref{fig:conflict_dynamics} further shows that after conflict is
introduced, the strongest baseline loses alignment while belief entropy
and persona deviation increase; CORE keeps alignment high and suppresses
both uncertainty and deviation from the anchored persona.
The largest improvements occur on ambiguity-positive and
conflict-positive turns, where uncontrolled updating is most brittle.

\paragraph{Boundary diagnostics.}
Because PERSIST is post-anchor by design, we separately evaluate
first-contact initialization and explicit preference reversal.
Under pressure-first initialization, CORE reduces FCR from \(31.6\%\)
for RLPA to \(20.9\%\), although this remains higher than CORE's
\(11.6\%\) FCR after anchoring, confirming that a grounded reference
belief strengthens conflict-aware preservation.
In preference-reversal cases, explicit and persistent changes can be
committed immediately, whereas tentative changes are deferred until
confirmation, showing that conflict induces caution rather than permanent
rejection.
Full posterior trajectories and turns to stable MAP are reported in
Appendix~\ref{app:boundary_diagnostics}.

\subsection{Mechanistic Ablations and Efficiency}
\label{sec:ablation}

\paragraph{Ablation Study.}
Table~\ref{tab:ablation} isolates the contribution of CORE.
Removing executable update actions causes the largest degradation:
AL drops from \(82.9\) to \(78.8\), CRR from \(82.7\) to \(71.6\),
and DRIFT increases from \(10.1\) to \(15.4\).
This pattern is consistent with executable routing contributing beyond
turn-local evidence extraction alone.
Other ablations weaken the same control pathway in different ways:
removing clarification mishandles under-resolved evidence, removing
inertia or optimized revision makes the posterior move too aggressively,
and removing calibration or revision rewards reduces conflict-sensitive
routing.

Failure-mode decomposition makes these effects more concrete:
removing executable update actions increases unsupported COMMIT from
\(11.3\%\) to \(22.9\%\), while removing clarification raises
under-clarification from \(18.4\%\) to \(42.7\%\).
Structure-matched controls provide a complementary test:
action prediction without executable belief revision reaches \(80.0\)
AL and \(72.1\) CRR, compared with \(82.9\) AL and \(82.7\) CRR for
CORE.
Together, these results support executable belief revision as a
contributing mechanism beyond intermediate traces, action-name
supervision, or additional structured labels alone.
Full controls and failure-mode breakdowns are reported in
Appendix~\ref{app:additional_results}.

\begin{table}[t]
\centering
\small
\begin{tabular}{lccc}
\toprule
Variant & AL$\uparrow$ & CRR$\uparrow$ & Drift$\downarrow$ \\
\midrule
CORE            & \textbf{82.9}$\pm$.4 & \textbf{82.7}$\pm$.4 & \textbf{10.1}$\pm$.4 \\
w/o update act. & 78.8$\pm$.5 & 71.6$\pm$.6 & 15.4$\pm$.5 \\
w/o clarify     & 80.1$\pm$.5 & 77.8$\pm$.5 & 13.9$\pm$.4 \\
w/o inertia     & 80.7$\pm$.4 & 78.6$\pm$.5 & 13.4$\pm$.4 \\
w/o opt. rev.   & 80.4$\pm$.5 & 78.1$\pm$.5 & 13.7$\pm$.5 \\
w/o cal. rew.   & 80.6$\pm$.4 & 76.4$\pm$.6 & 13.5$\pm$.4 \\
w/o rev. rew.   & 79.5$\pm$.5 & 75.2$\pm$.6 & 14.7$\pm$.5 \\
\bottomrule
\end{tabular}
\caption{\textbf{Mechanistic ablations.}
Removing components weakens alignment and stability.}
\label{tab:ablation}
\end{table}

\paragraph{Inference cost.}
CORE introduces a lightweight pre-generation control layer.
Figure~\ref{fig:4d} decomposes per-turn latency and shows that response
decoding remains the dominant cost, while update routing and belief
revision add only a small action-level overhead across
\textsc{Commit}, \textsc{Defer}, and \textsc{Ignore}.
The dialogue context is encoded once, slot-conditioned evidence logits
are produced in a shared pass, and cached slot/value descriptions avoid
separate LLM calls for each slot or value.
This yields \(O(KV_{\max}d)\) additional computation and adds only
23.0 ms per turn on the same backbone, suggesting that CORE improves
robustness without materially slowing inference.
Gold persona-state and update-action annotations are used only for
training or evaluation and are not required at inference time.
More details are given in
Appendix~\ref{app:latency_memory_retrieval}.

\paragraph{Human evaluation.}
Human-in-the-loop evaluation further examines evaluation reliability and
end-to-end interaction quality.
Across audited response, robustness, and pairwise items, human agreement
is strong, and judge predictions are well aligned with human judgments
(mean J--H \(=0.84\), mean \(\kappa=0.79\)), supporting the reliability
of response-level automatic evaluation.
In complete multi-turn interactions, CORE improves the average human
rating over RLPA by \(11.7\%\).
The gains are concentrated on persona consistency, ambiguity handling,
preference preservation, and long-horizon stability, which improve by
\(+11.6\%\)--\(15.5\%\), while response quality improves by only
\(+4.7\%\).
This indicates that CORE's advantage is concentrated more strongly on
long-horizon persona-state behavior than on generic response fluency.
This controlled evaluation provides behavioral validation under matched
multi-turn scenarios rather than evidence from natural longitudinal
deployment.
Details are in Appendix~\ref{app:evaluation_reliability}.

\section{Conclusion}

We presented CORE, which formulates long-horizon personalization as
controlled persona-state revision.
By separating turn-local evidence from persistent belief updates, CORE
preserves grounded preferences under unresolved observations while
remaining adaptive to sufficiently supported changes.
We also introduced PERSIST, a held-out benchmark for post-anchor
persona-state robustness under sequential interaction stress.
Across standard benchmarks, stress tests, and human evaluation, our
results support explicit update control as a useful mechanism for robust
long-horizon personalization.

\section*{Acknowledgements}
This work was supported in part by National Natural Science Foundation of China (62476070), the Major Key Project of PCL under Grant PCL2025A10 and PCL2024A06, Shenzhen Science and Technology Program \seqsplit{(RCJC20231211085918010, \, JCYJ20241202123503005, \, GXWD20231128103232001, \,ZDSYS20230626091203008,\, KQTD20240729102154066)}, Department of Science and Technology of Guangdong (2024A1515011540).

\section*{Limitations}

CORE relies on benchmark-instantiated persona slots and structured
supervision, so its state-level analysis does not fully capture
open-ended or cross-slot preferences.
Its conservative update policy may also delay adaptation when genuine
preference changes are weakly expressed.
Moreover, PERSIST primarily evaluates post-anchor robustness rather than
first-contact personalization, and controlled benchmarks cannot fully
represent long-term real-world interactions.

\section*{Ethical Considerations}

Persistent personalization raises privacy and user-control concerns
because maintained persona states can influence future interactions.
PERSIST is used only for controlled evaluation and is not intended for
profiling real users or inferring sensitive attributes.
Real-world systems should minimize retained information and allow users
to inspect, correct, delete, or opt out of persistent persona states.

\bibliography{custom}
\newpage
\onecolumn
\appendix
\twocolumn
\crefalias{section}{appendix}
\crefalias{subsection}{appendix}
\crefalias{subsubsection}{appendix}

\startcontents[appendix]

{
\hypersetup{linkcolor=black}
\printcontents[appendix]{ }{0}{\section*{Appendix}}
}

\newpage

\noindent\textbf{Appendix organization.}
The appendix provides details needed to audit and reproduce CORE.
Appendix~\ref{app:method_details} specifies the method implementation;
Appendix~\ref{app:experimental_training_protocol} reports experimental
controls, training setup, calibration, and PPO configuration;
Appendix~\ref{app:evaluation_reliability} defines the evaluation metrics
and reports judge and human-study reliability;
Appendix~\ref{app:persist} describes PERSIST construction, stress
strategies, and quality control; and
Appendix~\ref{app:additional_results} provides structure-matched controls,
boundary diagnostics, robustness breakdowns, efficiency analyses, and
failure-mode results.


\section{CORE Method Details}
\label{app:method_details}

\FloatBarrier

This section gives implementation details.

\subsection{Slot-Factorized Persona Belief}
\label{app:slot_belief}

CORE maintains a slot-factorized persona belief.
Let
\(\mathcal{K}=\{1,\ldots,K\}\)
denote the benchmark-instantiated slot inventory.

At turn \(t\), slot \(k\) has a working value set
\(\mathcal{V}_k^{(t)}\), and CORE stores a categorical posterior
\(b_t^k\in\mathcal{S}_k^{(t)}\), where
\(\mathcal{S}_k^{(t)}\) denotes the probability simplex over
\(\mathcal{V}_k^{(t)}\).

The joint belief is approximated as
\begin{equation}
    b_t(z)
    \approx
    \prod_{k=1}^{K} b^k_t(z^k),
    \qquad
    z^k \in \mathcal{V}^{(t)}_k .
    \label{eq:app_factorized_belief}
\end{equation}
This factorization is an auditable state interface, not an independence
assumption.
It lets each turn-level update be traced to a concrete persona slot.

For each slot, we summarize the posterior by its MAP value and anchored
confidence.
Let
\(\mathcal{V}^{(t)}_{k,+}
=
\mathcal{V}^{(t)}_k\setminus\{\mathrm{UNK}\}\)
and
\(m_t^k=1-b_t^k(\mathrm{UNK})\).
For \(m_t^k>0\), define
\(\bar b_t^k(v)=b_t^k(v)/m_t^k\)
over \(\mathcal{V}^{(t)}_{k,+}\).
Then
\begin{align}
    \hat{z}^{k}_{t}
    &=
    \arg\max_{v\in\mathcal{V}^{(t)}_k} b^k_t(v),
    \label{eq:app_slot_argmax}
    \\
    c_t^k
    &=
    m_t^k
    \begin{cases}
    1-\dfrac{H(\bar b_t^k)}{\log |\mathcal{V}^{(t)}_{k,+}|},
    & |\mathcal{V}^{(t)}_{k,+}|>1,\\[0.5ex]
    1, & |\mathcal{V}^{(t)}_{k,+}|=1,\\
    0, & |\mathcal{V}^{(t)}_{k,+}|=0.
    \end{cases}
    \label{eq:app_slot_confidence}
\end{align}
This prevents a posterior concentrated on \(\mathrm{UNK}\) from being
interpreted as strong persona grounding.

\subsection{Semi-Open Value Handling}
\label{app:semi_open_values}

The slot inventory is fixed within each benchmark, but value sets are
semi-open:
\begin{equation}
    \mathcal{V}^{(t)}_k
    =
    \mathcal{V}^{\mathrm{canon}}_k
    \cup
    \mathcal{V}^{\mathrm{obs}}_k(t)
    \cup
    \{\mathrm{UNK}\}.
    \label{eq:app_working_value_set}
\end{equation}
Here, \(\mathcal{V}^{\mathrm{canon}}_k\) contains benchmark-defined
canonical values,
\(\mathcal{V}^{\mathrm{obs}}_k(t)\) contains values observed or clarified
up to turn \(t\), and
\(\mathrm{UNK}\) marks unresolved evidence.
A new value is added only when it can be mapped to an existing slot with
sufficient semantic confidence.
Attributes without a reliable closed-slot mapping are assigned to a
residual open slot and excluded from closed-slot state metrics.

Surface mentions are normalized before entering
\(\mathcal{V}^{\mathrm{obs}}_k(t)\).
Paraphrases are merged only when they express the same persona commitment.
For example,
\textit{prefers concise answers} and
\textit{likes brief explanations}
may be merged, while
\textit{vegetarian} and \textit{vegan}
remain distinct unless the dialogue explicitly resolves them.
The same normalization rules are applied to all methods.

When a new value is admitted into
\(\mathcal{V}^{(t)}_k\),
the previous posterior \(b^k_{t-1}\) is extended with a
configuration-fixed prior mass and then renormalized over the expanded
support.
The semantic mapping used to determine whether a newly observed value
belongs to an existing slot is part of benchmark/schema normalization
rather than a separate inference-time slot-matching model.
Closed-slot belief metrics are computed only on normalized slots and
values.
Open-world attributes can still affect response-level evaluation, but
they are not counted as closed-slot belief correctness.

\subsection{Context Encoding and Slot-Wise Evidence Scoring}
\label{app:evidence_scoring}

At turn \(t\), CORE receives dialogue history \(H_{t-1}\),
user utterance \(u_t\), and previous belief \(b_{t-1}\).
A shared context encoder produces
\begin{equation}
    h_t
    =
    f_{\mathrm{enc}}(H_{t-1},u_t,b_{t-1}).
    \label{eq:app_context_encoder}
\end{equation}
The maintained belief is provided as a compact symbolic state summary,
including high-confidence slot values, unresolved \(\mathrm{UNK}\)
states, and historical state-control information available before the
current routing decision.
No current-turn update action is included in \(h_t\), because routing is
computed only after turn-local evidence and uncertainty descriptors have
been obtained.
This preserves the inference order
\(
\text{context encoding}
\rightarrow
\text{evidence scoring}
\rightarrow
\text{uncertainty characterization}
\rightarrow
\text{update routing}
\).

CORE does not use a separate inference-time slot matcher.
Each slot is represented by its schema, including a slot name, a short
natural-language definition, and the current candidate values in
\(\mathcal{V}^{(t)}_k\).
Slot grounding during inference is handled by slot-conditioned evidence
scoring, while value-to-slot normalization used to construct benchmark
targets is performed separately during preprocessing.

For each slot, we augment the working value set with a null option:
\begin{equation}
    \bar{\mathcal{V}}^{(t)}_k
    =
    \mathcal{V}^{(t)}_k \cup \{\bot\},
    \label{eq:app_augmented_value_set}
\end{equation}
where \(\bot\) means that the current turn provides no usable evidence
for slot \(k\).

For each \(v\in\bar{\mathcal{V}}^{(t)}_k\), CORE computes a local
evidence logit:
\begin{equation}
    \ell^k_t(v)
    =
    f_{\mathrm{evi}}(h_t,e_k,e_v).
    \label{eq:app_raw_evidence_score}
\end{equation}
The logit measures what the current turn locally suggests; it does not
decide whether the maintained belief should be revised.

For numerical stability, evidence logits are centered and clipped within
each slot:
\begin{align}
    \bar{\ell}^k_t(v)
    &=
    \ell^k_t(v)
    -
    \frac{1}{|\bar{\mathcal{V}}^{(t)}_k|}
    \sum_{v'\in\bar{\mathcal{V}}^{(t)}_k}
    \ell^k_t(v'),
    \label{eq:app_evidence_centering}
    \\
    \Delta^k_t(v)
    &=
    \mathrm{clip}
    \left(
        \frac{\bar{\ell}^k_t(v)}{s^k_t+\epsilon},
        -\Delta_{\max},
        \Delta_{\max}
    \right),
    \label{eq:app_evidence_clipping}
\end{align}
where \(s^k_t\) is the within-slot scale used by the implementation and
\(\Delta_{\max}\) is fixed before evaluation.
Both quantities are shared across experiments rather than selected from
test performance.
For candidate sets for which scale estimation is unreliable, the
implementation applies clipping without the scale-normalization step.
In the rest of the appendix,
\(\Delta^k_t(v)\) denotes this stabilized evidence score, matching the
evidence notation in the main text.

The stabilized scores define a turn-local distribution over the augmented
value space:
\begin{equation}
    \widetilde{p}^{k}_{t}(v)
    =
    \frac{
        \exp(\Delta^k_t(v))
    }{
        \sum_{v'\in\bar{\mathcal{V}}^{(t)}_k}
        \exp(\Delta^k_t(v'))
    },
    \qquad
    v\in\bar{\mathcal{V}}^{(t)}_k .
    \label{eq:app_turn_local_distribution}
\end{equation}
This distribution is turn-local.
It identifies which value, if any, is supported by the current
observation; routing decides whether that evidence should be committed,
deferred, or ignored.

\subsection{Uncertainty-Guided Update Routing}
\label{app:update_routing}

CORE derives relevance, ambiguity, and conflict from the turn-local
evidence distribution.
The null option gives the relevance estimate:
\begin{equation}
    r^k_t
    =
    1-\widetilde{p}^{k}_{t}(\bot).
    \label{eq:app_relevance}
\end{equation}

For non-null values, we renormalize the evidence distribution over
\(\mathcal{V}^{(t)}_k\):
\begin{equation}
    \widetilde{p}^{k}_{t,+}(v)
    =
    \frac{
        \widetilde{p}^{k}_{t}(v)
    }{
        \sum_{v'\in\mathcal{V}^{(t)}_k}
        \widetilde{p}^{k}_{t}(v')
    },
    \qquad
    v\in\mathcal{V}^{(t)}_k .
    \label{eq:app_non_null_distribution}
\end{equation}

Ambiguity is computed as
\begin{equation}
    a^k_t
    =
    \begin{cases}
    \dfrac{H(\widetilde{p}^{k}_{t,+})}
    {\log |\mathcal{V}^{(t)}_k|},
    & |\mathcal{V}^{(t)}_k|>1,\\[0.8ex]
    0, & |\mathcal{V}^{(t)}_k|\le 1.
    \end{cases}
    \label{eq:app_ambiguity}
\end{equation}

When a new value expands
\(\mathcal{V}^{(t)}_k\),
we first extend \(b^k_{t-1}\) with the same configuration-fixed prior
mass used in Appendix~\ref{app:semi_open_values} and renormalize it over
the aligned support.
Conflict is then
\begin{equation}
    q^k_t
    =
    r^k_t
    D_{\mathrm{JS}}
    \left(
        \widetilde{p}^{k}_{t,+}
        \,\|\,
        b^k_{t-1}
    \right).
    \label{eq:app_ambiguity_conflict}
\end{equation}
The multiplier \(r^k_t\) prevents weakly matched turns from being treated
as strong conflicts.
When \(r^k_t\) falls below the calibrated relevance threshold, slot
\(k\) is treated as unmatched and bypasses persistent revision.

The update policy directly conditions on the current evidence distribution,
previous belief, and uncertainty descriptors:
\begin{equation}
\begin{gathered}
    \alpha_t^k
    \sim
    \pi_{\mathrm{upd}}^\theta
    \Big(
    \alpha \,\big|\,
    \widetilde{p}^{k}_{t},
    b^k_{t-1},
    r^k_t,
    a^k_t,
    q^k_t,
    c^k_{t-1}
    \Big),\\[2pt]
    \alpha_t^k
    \in
    \{\mathrm{COMMIT},\mathrm{DEFER},\mathrm{IGNORE}\}.
\end{gathered}
\label{eq:app_update_policy}
\end{equation}
The implementation uses the corresponding slot-conditioned encoded
representation, avoiding a literal flattening of variable-length candidate
distributions while preserving the routing semantics in the main text.

The action semantics are:
\begin{itemize}[leftmargin=1.5em,itemsep=1pt,topsep=2pt]
    \item \textbf{COMMIT}: authorize sufficiently grounded evidence to
    revise the maintained belief.
    \item \textbf{DEFER}: preserve the current belief while exposing
    relevant but under-resolved evidence to the current-turn discourse
    policy.
    \item \textbf{IGNORE}: exclude unrelated or explicitly non-persistent
    evidence from the active update pathway.
\end{itemize}

DEFER does not create a separate durable pending-memory buffer.
Deferred evidence remains current-turn control information and does not
automatically become a future COMMIT.
If related evidence or clarification later appears while the earlier
dialogue context remains accessible, the slot is inferred and routed
again from the newly available context.
DEFER is an internal update action, whereas CLARIFY is a dialogue-level
decision made after slot-level revision.

\subsection{Belief Revision}
\label{app:belief_revision}

After routing, CORE converts each update decision into a continuous gate.
Following Section~\ref{method_3.3}, the gate feature is
\begin{equation}
    \psi^k_t
    =
    [
        r^k_t;
        1-a^k_t;
        1-q^k_t;
        c^k_{t-1}
    ].
    \label{eq:app_gate_features}
\end{equation}

The revision gate is
\begin{equation}
    g^k_t
    =
    \mathbb{I}[\alpha^k_t=\mathrm{COMMIT}]
    \cdot
    \sigma(w_g^\top\psi^k_t+b_g).
    \label{eq:app_gate_action_conditioned}
\end{equation}
DEFER and IGNORE set \(g^k_t=0\).
Under COMMIT, the update magnitude is conditioned on relevance,
ambiguity, conflict, and prior anchored confidence; we do not assume a
fixed monotonic effect for any individual descriptor unless imposed by
the implementation.

Given the gate, CORE performs a KL-regularized posterior update over the
non-null value set \(\mathcal{V}^{(t)}_k\):
\begin{equation}
\scalebox{0.85}{
$\displaystyle
b^k_t
=
\arg\min_{p\in\Delta(\mathcal{V}^{(t)}_k)}
\left[
    \mathrm{KL}(p\|b^k_{t-1})
    -
    g^k_t
    \sum_{v\in\mathcal{V}^{(t)}_k}
    p(v)\Delta^k_t(v)
\right].
$}
\label{eq:app_kl_revision}
\end{equation}

The null option \(\bot\) is used only for evidence relevance and is never
written into the persona belief.
The solution is
\begin{equation}
\begin{gathered}
    b^k_t(v)
    =
    \frac{
        b^k_{t-1}(v)
        \exp(g^k_t\Delta^k_t(v))
    }{
        \sum_{v'\in\mathcal{V}^{(t)}_k}
        b^k_{t-1}(v')
        \exp(g^k_t\Delta^k_t(v'))
    }, \\[1ex]
    v\in\mathcal{V}^{(t)}_k .
\end{gathered}
\label{eq:app_closed_form_update}
\end{equation}

Thus \(g^k_t=0\) preserves the prior, while larger \(g^k_t\) moves the
posterior toward evidence-supported values.
For DEFER and IGNORE, the posterior remains unchanged, and deferred
evidence is used only for current-turn discourse control.

After slot-level revision, CORE decides whether to respond or request
clarification:
\begin{equation}
\begin{aligned}
d_t \sim \pi^\theta_{\mathrm{dis}}
\Big(
d \,\big|\,
&u_t,\,H_{t-1},\,b_t,\,
\{r_t^k,a_t^k,q_t^k,\alpha_t^k\}_{k=1}^{K}
\Big),\\
&d_t\in\{\mathrm{RESPOND},\mathrm{CLARIFY}\}.
\end{aligned}
\label{eq:app_dialogue_decision}
\end{equation}
A deferred observation does not automatically trigger clarification.

\subsection{Inference-Time Algorithm}
\label{app:inference_algorithm}

Algorithm~\ref{alg:core_inference} summarizes inference.
Evidence is scored before routing; only evidence routed through COMMIT
can revise the maintained belief.

\begin{algorithm}[t]
\caption{CORE inference-time algorithm}
\label{alg:core_inference}
\small
\begin{algorithmic}[1]
\Require History $H_{t-1}$, user utterance $u_t$, previous belief $b_{t-1}$
\Ensure Updated belief $b_t$ and response $y_t$

\State $h_t \leftarrow f_{\mathrm{enc}}(H_{t-1},u_t,b_{t-1})$
\State $\mathcal{A}_{\mathrm{upd}}\leftarrow\{\mathrm{COMMIT},\mathrm{DEFER},\mathrm{IGNORE}\}$
\State $\mathcal{A}_{\mathrm{dis}}\leftarrow\{\mathrm{RESPOND},\mathrm{CLARIFY}\}$

\For{$k=1,\ldots,K$}
    \State Construct $\mathcal{V}^{(t)}_k$
    \State $\bar{\mathcal{V}}^{(t)}_k\leftarrow\mathcal{V}^{(t)}_k\cup\{\bot\}$
    \State Score $\ell^k_t(v)$ for all $v\in\bar{\mathcal{V}}^{(t)}_k$
    \State Stabilize scores to obtain $\Delta^k_t(v)$
    \State Compute $\widetilde{p}^{k}_{t}$ over $\bar{\mathcal{V}}^{(t)}_k$
    \State $r^k_t\leftarrow1-\widetilde{p}^{k}_{t}(\bot)$

    \If{$r^k_t$ is below the relevance threshold}
        \State $\alpha^k_t\leftarrow\mathrm{IGNORE}$
        \State $b^k_t\leftarrow b^k_{t-1}$
    \Else
        \State Compute $\widetilde{p}^{k}_{t,+}$ over $\mathcal{V}^{(t)}_k$
        \State Compute $a^k_t$ and $q^k_t$ using Eq.~\eqref{eq:app_ambiguity_conflict}
        \State Predict $\alpha_t^k$ using Eq.~\eqref{eq:app_update_policy}

        \If{$\alpha^k_t=\mathrm{COMMIT}$}
            \State Compute $g^k_t$ using Eq.~\eqref{eq:app_gate_action_conditioned}
            \State Update $b^k_t$ using Eq.~\eqref{eq:app_closed_form_update}
        \Else
            \State $b^k_t\leftarrow b^k_{t-1}$
        \EndIf
    \EndIf
\EndFor

\State Predict $d_t$ using Eq.~\eqref{eq:app_dialogue_decision}
\State $y_t\sim\pi^\theta_{\mathrm{resp}}(\cdot\mid H_{t-1},u_t,b_t,d_t)$
\State \Return $b_t,y_t$
\end{algorithmic}
\end{algorithm}

\paragraph{Gradient boundaries.}
During supervised warm-up, gradients are applied to the evidence scorer,
update router, discourse policy, and response policy through their
corresponding supervised losses. The belief update in
Eq.~\eqref{eq:app_closed_form_update} remains differentiable with respect
to the evidence scores and revision gate, while the discrete routing action
is optimized through the routing loss and policy objective.


\section{Experimental and Training Protocol}
\label{app:experimental_training_protocol}

This section specifies the experimental and training protocol behind
Section~\ref{sec:experiments}.
It covers benchmark splits, slot instantiation, baseline controls,
update-action labels, label auditing, supervised warm-up, uncertainty
calibration, PPO rewards, and stability checks.
The goal is to make the comparison auditable and to test whether the
observed gains can be distinguished from alternative explanations such
as hidden supervision, larger context budgets, reward artifacts, or
degenerate conservatism.

\subsection{Benchmarks, Splits, and Evaluation Scope}
\label{app:benchmarks_splits}

All methods are evaluated in the same partially observed interaction
setting.
The model observes the dialogue history and current user utterance, but
never observes the latent persona state directly.
It must infer, maintain, and update personalization from turn-level
evidence.

We use three complementary benchmarks.
ALOE evaluates progressive preference revelation;
PersonaChat evaluates generalization to free-form persona attributes;
and PERSIST evaluates robustness under ambiguity, conflict, and
controlled social influence.
For ALOE and PersonaChat, splits are constructed at the persona level
whenever profile identifiers are available.
For datasets without explicit identifiers, we cluster normalized persona
descriptors and split clusters rather than individual turns.

Table~\ref{tab:app_benchmark_splits} reports the processed splits for
ALOE and PersonaChat.
The calibration split is used for checkpoint selection, threshold
calibration, response-judge calibration, and uncertainty calibration
when applicable.
PERSIST is used only for held-out robustness evaluation; its benchmark
construction and evaluation protocol are described separately in
Appendix~D.

\begin{table}[!t]
\centering
\small
\setlength{\tabcolsep}{4pt}
\renewcommand{\arraystretch}{1.08}
\caption{\textbf{Processed experimental splits.}
ALOE and PersonaChat splits are fixed before training and shared by all
methods.}
\label{tab:app_benchmark_splits}
\begin{tabularx}{\linewidth}{@{}lrrrrX@{}}
\toprule
\textbf{Benchmark}
& \textbf{Train}
& \textbf{Cal.}
& \textbf{Test}
& \textbf{Turns}
& \textbf{Split guarantee} \\
\midrule
ALOE
& 4,800 & 600 & 600 & 61,248
& preference-cluster disjoint \\
PersonaChat
& 8,448 & 1,024 & 1,024 & 103,872
& persona-profile disjoint \\
\bottomrule
\end{tabularx}
\end{table}

Response-level metrics are computed on all retained examples.
Closed-slot belief metrics are computed only when the relevant persona
evidence can be reliably mapped to a normalized slot-value target.
Residual open-world attributes remain in response-level evaluation but
are excluded from closed-slot state metrics.

\subsection{Slot Instantiation and Baseline Controls}
\label{app:slot_baseline_controls}

CORE uses the same state-control interface across benchmarks, while the
slot inventory is benchmark-instantiated.
This avoids imposing a universal ontology while keeping evaluation
auditable.
ALOE, PersonaChat, and PERSIST instantiate 10, 8, and 12 closed slots,
respectively.
Their closed-slot coverage after normalization is 91.4\%, 86.6\%, and
89.1\%; the remaining attributes are assigned to residual open slots and
excluded only from closed-slot state metrics.
Typical slots cover explanation style, privacy, budget, lifestyle, diet,
verbosity, formality, travel, and risk tolerance.

Surface mentions are normalized before evaluation.
Paraphrases are merged only when they express the same stable commitment.
For example,
\textit{prefers concise answers}
and
\textit{likes brief explanations}
may share a verbosity value, while
\textit{vegetarian}
and
\textit{vegan}
remain distinct unless explicitly clarified.
The same normalization rules are applied to CORE and all baselines.

For systems that do not expose a native structured persona belief,
state-level diagnostics are computed with the same frozen evaluation-time
state extractor, applied uniformly across those systems. These quantities
should be interpreted as evaluation-time diagnostics of observable
persona-state content rather than as claims about a baseline's native
internal representation.

Baselines test six alternative explanations:
prompting or reasoning traces
(Reminder, Self-Critic, CoT),
retrieved or persistent memory
(RAG top-$5$, MemoryBank),
offline preference optimization
(SFT, DPO),
personalized-agent control
(RLPA),
intermediate-format supervision
(structure-matched controls),
and scale/backbone effects.
Same-backbone methods use matched decoding and calibration protocols;
structure-matched controls are reported in
Appendix~\ref{app:additional_results}.

All baselines use the same dialogue input and context budget.
Reminder prepends a persona summary,
Self-Critic revises drafts with persona-consistency critique,
CoT adds reasoning instructions,
and RAG retrieves top-5 history snippets.
MemoryBank maintains persistent textual memory, while RLPA keeps its
personalized-agent interface without CORE-style belief revision.
SFT and DPO use the same processed trajectories with matched
optimization and checkpoint selection.

All same-backbone methods use Llama-3.2-3B-Instruct with an 8k context,
the same processed training pool, matched optimization and checkpoint
selection, and 256-token decoding at temperature 0.7; retrieval baselines
use top-$5$ retrieval unless otherwise stated.

CORE's compact belief summary counts against the same context budget
used by memory and retrieval baselines.
Retrieval baselines may access raw retrieved histories, whereas CORE
uses compressed belief states and compact historical state-control summaries.
Thus, the methods share the same context budget while differing in how
persona information is represented and maintained.

Gold normalized persona states and update-action annotations are used
for supervision or evaluation only.
At inference time, CORE operates from dialogue context, its maintained
belief, predicted turn-local evidence, and predicted routing decisions;
no gold state or action annotation is required.

\subsection{Update-Action Labels and Audit}
\label{app:update_label_audit}

CORE uses slot-turn update labels for supervised warm-up on ALOE and
PersonaChat and for decision-level evaluation.
Corresponding PERSIST labels are diagnostic only and are never used for
CORE training or calibration.
These labels are constructed independently of CORE predictions.
Each label belongs to
\begin{equation}
\alpha^{k,\star}_t
\in
\{\text{COMMIT},\text{DEFER},\text{IGNORE}\}.
\label{eq:app_update_label_space}
\end{equation}

The label rule uses four observable factors:
\(\mathrm{Rel}_t^k\) indicates whether turn \(t\) contains evidence
about slot \(k\);
\(\mathrm{Amb}_t^k\) indicates whether the relevant evidence remains
under-resolved;
\(\mathrm{UConf}_t^k\) indicates whether the evidence conflicts with a
grounded prior while the conflict remains unresolved; and
\(\mathrm{Stable}_t^k\) indicates whether the evidence expresses a
persistent rather than explicitly local commitment.

Here, \(\mathrm{Stable}^{k}_t=0\) is reserved for evidence that is
explicitly task-local, transient, hypothetical, quoted from another
speaker, or otherwise not intended to characterize the user's persistent
persona. Uncertainty about a potentially persistent preference is instead
handled by DEFER.

Candidate labels are assigned by the mutually exclusive rule
\begin{equation}
\alpha^{k,\star}_t=
\begin{cases}
\mathrm{IGNORE},
&
\mathrm{Rel}^{k}_t=0
\;\lor\;
\mathrm{Stable}^{k}_t=0,
\\[0.6ex]
\mathrm{DEFER},
&
\mathrm{Rel}^{k}_t=1,\;
\mathrm{Stable}^{k}_t=1,
\\
&
\mathrm{Amb}^{k}_t=1
\;\lor\;
\mathrm{UConf}^{k}_t=1,
\\[0.6ex]
\mathrm{COMMIT},
&
\mathrm{Rel}^{k}_t=1,\;
\mathrm{Stable}^{k}_t=1,
\\
&
\mathrm{Amb}^{k}_t=0,\;
\mathrm{UConf}^{k}_t=0.
\end{cases}
\label{eq:app_update_label_rule}
\end{equation}

COMMIT denotes sufficiently resolved evidence for persistent revision.
DEFER denotes relevant evidence that may be persistent but whose value or
relation to the grounded prior remains unresolved.
IGNORE denotes either unrelated evidence or evidence whose semantics are
explicitly local or non-persistent.
The three categories are mutually exclusive by construction.

An observed conflict does not imply permanent rejection.
If subsequent evidence resolves the conflict and clearly establishes a
persistent preference change, the new observation is re-evaluated and
may receive COMMIT.

We audit labels in four stages.
First, a proposer constructs a candidate update label from the dialogue
context, target slot, previous belief summary, current utterance, and
benchmark metadata.
Second, an independent verifier assigns COMMIT, DEFER, or IGNORE from the
same observable dialogue information.
The verifier is not shown the proposer's factor annotations, proposed
action, confidence, rationale, or any CORE prediction.
Third, intent-drift filtering removes examples whose realized evidence
type does not match the intended condition.
Finally, a stratified human audit checks slot mapping, action semantics,
and false-COMMIT risk.
All filtering and auditing are performed before CORE training and never
depend on CORE success or failure.

Table~\ref{tab:app_update_label_audit_summary} merges label distribution
and audit statistics.

\begin{table*}[!t]
\centering
\small
\setlength{\tabcolsep}{5pt}
\renewcommand{\arraystretch}{1.08}
\caption{\textbf{Update-label construction and audit summary.}
Labels are built before training and filtered independently of CORE
predictions.}
\label{tab:app_update_label_audit_summary}
\begin{tabularx}{\textwidth}{@{}lrrrrrrX@{}}
\toprule
\textbf{Benchmark}
& \textbf{COMMIT}
& \textbf{DEFER}
& \textbf{IGNORE}
& \textbf{Verifier agree}
& \textbf{Drift removed}
& \textbf{Final retained}
& \textbf{Main removed pattern} \\
\midrule
ALOE
& 43.8 & 24.6 & 31.6 & 89.6\% & 6.1\% & 157,384
& Local task requests mislabeled as stable preferences. \\
PersonaChat
& 39.5 & 27.2 & 33.3 & 86.8\% & 7.4\% & 198,912
& Free-form persona statements with underspecified slot mapping. \\
PERSIST
& 31.6 & 32.4 & 36.0 & 84.9\% & 9.8\% & 247,536
& Stress turns whose pressure was weaker than intended. \\
\bottomrule
\end{tabularx}
\end{table*}

Verifier--human agreement is strongest for COMMIT, while most residual
disagreement occurs between DEFER and IGNORE; this mainly affects fine-grained
ACTACC because both actions avoid unsupported commitment on conflict-positive
turns.

\subsection{Supervised Warm-Up and Uncertainty Calibration}
\label{app:supervised_calibration}

We use the supervised objective defined in Section~\ref{method_3.4}.
The evidence target is \(\bot\) when the current turn provides no usable
evidence for slot \(k\). After warm-up, calibration parameters associated
with relevance, ambiguity, and conflict are selected on the held-out
calibration split and fixed during PPO. This prevents policy optimization
from improving reward by reshaping the uncertainty estimates themselves.

\subsection{PPO Configuration and Stability Checks}
\label{app:ppo_reward_design}

All same-backbone trainable methods use Llama-3.2-3B-Instruct with the
same tokenizer.
Experiments are run on NVIDIA A800 80GB GPUs.
All trainable methods are implemented with PyTorch and Hugging Face
Transformers using Llama-3.2-3B-Instruct and its default tokenizer.
RAG uses top-$5$ retrieval, and all generation-based methods use matched
decoding settings.
All reported results use matched training pools, effective batch sizes,
update steps, decoding settings, and checkpoint-selection protocols.

PPO starts from the supervised warm-up checkpoint
\(\pi_{\mathrm{SFT}}\).
The policy state contains the dialogue representation, maintained belief
summary, calibrated uncertainty descriptors, current-turn routing descriptors,
previous update actions, and the current discourse decision.
It does not contain gold slot labels or gold update actions at inference
time.

We optimize
\begin{equation}
\begin{aligned}
    \mathcal{L}_{\mathrm{PPO}}
    ={}&
    -\mathbb{E}_t
    \Big[
    \min
    \big(
    \rho_tA_t,
    \mathrm{clip}(\rho_t,1-\epsilon,1+\epsilon)A_t
    \big)
    \Big]
    \\
    &+
    \beta_{\mathrm{KL}}
    D_{\mathrm{KL}}
    \big(
    \pi_\theta(\cdot|s_t)
    \,\|\,
    \pi_{\mathrm{SFT}}(\cdot|s_t)
    \big)
    \\
    &+
    \lambda_V\mathcal{L}_V
    -
    \lambda_H\mathcal{H}(\pi_\theta),
\end{aligned}
\label{eq:app_ppo_objective}
\end{equation}
where
\(\rho_t\)
is the policy ratio,
\(A_t\)
is the estimated advantage,
\(\mathcal{L}_V\)
is the value loss, and
\(\mathcal{H}\)
is policy entropy.

The task reward is
\begin{equation}
\begin{split}
    R_t
    ={}&
    \lambda_{\mathrm{act}}R^{\mathrm{act}}_t
    +
    \lambda_{\mathrm{state}}R^{\mathrm{state}}_t
    \\
    &+
    \lambda_{\mathrm{ans}}R^{\mathrm{ans}}_t
    +
    \lambda_{\mathrm{clar}}R^{\mathrm{clar}}_t ,
\end{split}
\label{eq:app_expanded_reward}
\end{equation}
while the KL regularizer is applied separately in
Eq.~\eqref{eq:app_ppo_objective}.

We set the reward weights for action correctness, state fidelity,
response quality, and clarification behavior to \(0.35\), \(0.30\),
\(0.25\), and \(0.10\), respectively.

The reward components correspond to distinct training signals.
\(R^{\mathrm{act}}_t\)
evaluates COMMIT, DEFER, and IGNORE routing;
\(R^{\mathrm{state}}_t\)
evaluates fidelity of the maintained persona belief;
\(R^{\mathrm{ans}}_t\)
evaluates personalized response quality; and
\(R^{\mathrm{clar}}_t\)
evaluates whether clarification is used when relevant evidence remains
unresolved while penalizing unnecessary clarification on well-resolved
turns.

\begin{table}[!t]
\centering
\small
\setlength{\tabcolsep}{4pt}
\renewcommand{\arraystretch}{1.08}
\caption{\textbf{PPO configuration.}
The same setting is used across CORE PPO runs unless otherwise stated.}
\label{tab:app_ppo_config}
\begin{tabularx}{\linewidth}{@{}p{0.38\linewidth}X@{}}
\toprule
\textbf{Hyperparameter} & \textbf{Value} \\
\midrule
Reference policy & Frozen SFT checkpoint \\
PPO clip range $\epsilon$ & 0.20 \\
KL coefficient $\beta_{\mathrm{KL}}$ & 0.05 \\
Value-loss weight $\lambda_V$ & 0.50 \\
Entropy weight $\lambda_H$ & 0.01 \\
GAE $\lambda$ & 0.95 \\
Discount factor $\gamma$ & 0.99 \\
Effective batch size & 128 \\
PPO epochs / batch & 4 \\
Max gradient norm & 1.0 \\
\bottomrule
\end{tabularx}
\end{table}

We track both optimization-level and behavior-level diagnostics during
PPO.
Optimization diagnostics include total reward, component rewards, KL
divergence, policy entropy, clip fraction, value loss, and gradient norm.
Behavior diagnostics include action frequencies, clarification rate under
ambiguity, false commitment, missed update rate, and belief drift.

A robust policy should not obtain high apparent safety by always
deferring, always ignoring, or always asking clarification.
We therefore audit collapse to always-DEFER or always-IGNORE, excessive
clarification on low-ambiguity turns, verbose but weakly informative
personalization, correct-sounding responses paired with inaccurate belief
updates, and superficial style mimicry that does not improve decision
quality.
Detailed reward-hacking and degeneracy diagnostics are reported in
Appendix~\ref{app:additional_results}.

\FloatBarrier


\section{Evaluation Metrics, Reliability, and Human Study}
\label{app:evaluation_reliability}

This section defines the metrics used in
Section~\ref{sec:experiments}
and reports reliability checks for both automatic and human evaluation.
The central principle is
\emph{metric-source separation}:
generated responses, maintained beliefs, update actions, and clarification
decisions are evaluated from different objects and supervision sources.
This prevents response-level judge scores from determining state- or
action-level conclusions.

\subsection{Metric-Source Separation}
\label{app:metric_source_separation}

We evaluate generated responses, maintained beliefs, update actions, and
clarification decisions from distinct sources. GPT-5.1 is used for AL and
ARUS and for an auxiliary pairwise reliability audit. Belief- and
decision-level metrics use reference annotations, maintained states, logged
native actions, or the fixed behavior-derived proxies in
Appendix~\ref{app:decision_metrics}. Response-level outcomes and human
evaluation are treated as primary end-to-end evidence; state and action
metrics are complementary diagnostics.

\subsection{Response-Level Metrics}
\label{app:response_metrics}

\paragraph{Alignment Level.}
Alignment Level (AL) is the primary response-level metric.
For dialogue \(j\) at turn \(t\), the judge assigns
\begin{equation}
    s_{j,t}\in\{1,2,3,4,5\},
    \label{eq:app_al_score}
\end{equation}
where higher scores indicate stronger personalized alignment, better use
of grounded persona evidence, and fewer violations of established
commitments.
We report AL on a 0--100 scale:
\begin{equation}
\begin{aligned}
\mathrm{AL}_t
&=
100\cdot
\frac{
\frac{1}{|\mathcal{J}_t|}
\sum_{j\in\mathcal{J}_t}s_{j,t}
-1
}{4},
\\[-1mm]
\mathrm{AL}
&=
\frac{1}{T}
\sum_{t=1}^{T}
\mathrm{AL}_t .
\end{aligned}
\label{eq:app_al}
\end{equation}
Because the underlying judge score is ordinal, we interpret AL together
with pairwise preference and human-agreement checks.

\paragraph{Normalized Improvement Rate.}
We normalize turn-level AL by
\begin{equation}
    \widetilde{\mathrm{AL}}_t
    =
    \frac{\mathrm{AL}_t}{100}.
    \label{eq:app_al_norm}
\end{equation}
We then fit
\begin{equation}
    \widetilde{\mathrm{AL}}_t
    =
    \beta_0+\beta_1t+\epsilon_t,
    \label{eq:app_nir_regression}
\end{equation}
and define
\begin{equation}
    \mathrm{N\text{-}IR}
    =
    \beta_1.
    \label{eq:app_nir}
\end{equation}
Higher N-IR means that the model uses accumulating persona evidence more
efficiently.

\paragraph{Trajectory regularity.}
Normalized \(R^2\) measures whether alignment improves smoothly:
\begin{equation}
    \mathrm{N\text{-}R}^2
    =
    1-
    \frac{
        \sum_{t=1}^{T}
        \left(
            \widetilde{\mathrm{AL}}_t
            -
            \widehat{\widetilde{\mathrm{AL}}}_t
        \right)^2
    }{
        \sum_{t=1}^{T}
        \left(
            \widetilde{\mathrm{AL}}_t
            -
            \overline{\widetilde{\mathrm{AL}}}
        \right)^2
    }.
    \label{eq:app_nr2}
\end{equation}

\subsection{Belief-Level Metrics}
\label{app:belief_metrics}

Belief-level metrics evaluate the maintained persona state rather than
the generated response.
They are computed only on closed-slot examples with reliable gold
slot-value annotations.
Let
\(z^{k,\star}_{j,t}\)
be the gold value for slot \(k\) in dialogue \(j\) at turn \(t\), and
let
\(b^k_{j,t}\)
be the model's posterior belief.

\paragraph{Slot accuracy.}
\begin{equation}
    \mathrm{SLOTACC}
    =
    \frac{100}{|\mathcal{S}|}
    \sum_{(j,t,k)\in\mathcal{S}}
    \mathbb{I}
    \left[
        \operatorname*{argmax}_{v}
        b^k_{j,t}(v)
        =
        z^{k,\star}_{j,t}
    \right].
    \label{eq:app_slotacc}
\end{equation}

\paragraph{Belief update accuracy.}
Let \(\mathcal{U}\) be the set of update-positive slot-turns where the
gold state should change or be confirmed by reliable evidence.
Then
\begin{equation}
    \mathrm{BUA}
    =
    100\cdot
    \frac{1}{|\mathcal{U}|}
    \sum_{(j,t,k)\in\mathcal{U}}
    \mathbb{I}
    \left[
        \arg\max_v b^k_{j,t}(v)
        =
        z^{k,\star}_{j,t}
    \right].
    \label{eq:app_bua}
\end{equation}
BUA checks whether a model can adapt when evidence is reliable, so high
robustness cannot be explained by complete inertia.

\paragraph{Persona-state drift.}
\begin{equation}
    \mathrm{DRIFT}
    =
    100\cdot
    \frac{1}{|\mathcal{S}|}
    \sum_{(j,t,k)\in\mathcal{S}}
    \left(
        1-b^k_{j,t}(z^{k,\star}_{j,t})
    \right).
    \label{eq:app_drift}
\end{equation}
Lower DRIFT means less posterior mass is assigned away from the gold
persona value.

\paragraph{Belief calibration.}
We additionally assess whether slot confidence tracks empirical MAP
correctness through the reliability analysis reported with Figure~3(c) in the
main text.

\subsection{Decision-Level Metrics}
\label{app:decision_metrics}

Decision-level metrics are mechanism diagnostics for update control
rather than primary measures of end-to-end system quality.
For CORE, they are computed from logged native update actions and
reference labels.
For systems without explicit update actions, a fixed behavior-derived
proxy is constructed from observable post-turn behavior:
persistent adoption of the new evidence is mapped to COMMIT,
explicit unresolved handling or clarification to DEFER,
and preservation of the prior commitment without incorporating the new
cue to IGNORE.
These proxies are used only for diagnostic action metrics, are not used
for training or model selection, and should not be interpreted as native
internal actions of the baseline systems.

\paragraph{Action accuracy.}
\begin{equation}
    \mathrm{ACTACC}
    =
    100\cdot
    \frac{1}{|\mathcal{A}|}
    \sum_{(j,t,k)\in\mathcal{A}}
    \mathbb{I}
    [
        \alpha^k_{j,t}
        =
        \alpha^{k,\star}_{j,t}
    ].
    \label{eq:app_actacc}
\end{equation}

\paragraph{Conflict non-commitment rate (CRR).}
Let \(\mathcal{C}\) denote conflict-positive slot-turns where an
unsupported persistent commitment should be avoided.
CRR measures whether the model avoids COMMIT at this level; it does not
attempt to distinguish the finer DEFER-versus-IGNORE boundary.
\begin{equation}
\mathrm{CRR}
=
\frac{100}{|\mathcal{C}|}
\sum_{(j,t,k)\in\mathcal{C}}
\mathbb{I}
\left[
\alpha^k_{j,t}
\in
\{\textsc{Defer},\textsc{Ignore}\}
\right].
\label{eq:app_crr}
\end{equation}

\paragraph{False commitment rate.}
Let \(\mathcal{N}\) be ambiguity-positive or conflict-positive slot-turns
whose reference action is not COMMIT:
\begin{equation}
    \mathrm{FCR}
    =
    100\cdot
    \frac{1}{|\mathcal{N}|}
    \sum_{(j,t,k)\in\mathcal{N}}
    \mathbb{I}
    [
        \alpha^k_{j,t}
        =
        \mathrm{COMMIT}
    ].
    \label{eq:app_fcr}
\end{equation}

\paragraph{Adaptive probing rate.}
Let \(\mathcal{Q}\) be ambiguity-positive turns for which clarification
is preferred:
\begin{equation}
    \mathrm{APR}
    =
    100\cdot
    \frac{1}{|\mathcal{Q}|}
    \sum_{(j,t)\in\mathcal{Q}}
    \mathbb{I}
    [
        d_{j,t}
        =
        \mathrm{CLARIFY}
    ].
    \label{eq:app_apr}
\end{equation}

\subsection{PERSIST Robustness Metrics}
\label{app:persist_robustness_metrics}

PERSIST evaluates whether a model preserves grounded persona commitments
under ambiguity, conflict, and controlled social influence.
It does not require models to expose CORE-specific actions.

\paragraph{Axiom Recall Under Stress.}
For confrontation turns with stressed axiom \(a_j\), ARUS measures
whether the response preserves the anchored axiom:
\begin{equation}
\mathrm{ARUS}
=
\frac{100}{|\mathcal{P}|}
\sum_{j\in\mathcal{P}}
\mathbb{I}
\left[
\mathrm{JudgePreserve}(y_j,a_j)=1
\right].
\label{eq:app_arus}
\end{equation}

\paragraph{Blind Compliance Defense.}
Let \(\mathcal{B}\) denote pressure-positive turns where blind compliance
would overwrite grounded persona evidence.
BCD is
\begin{equation}
    \mathrm{BCD}
    =
    100\cdot
    \frac{1}{|\mathcal{B}|}
    \sum_{j\in\mathcal{B}}
    \mathbb{I}
    [
        y_j
        \in
        \mathcal{Y}^{\mathrm{acceptable}}_j
    ].
    \label{eq:app_bcd}
\end{equation}
Acceptable behavior includes preserving the anchored commitment,
resisting unsupported pressure, or responding cautiously.
The acceptable-behavior set is specified by PERSIST's behavior-level
target annotations before model evaluation, so the target set itself is
independent of any evaluated system prediction.

\paragraph{Strategy-level macro average.}
For strategy \(s\), let \(\mathcal{D}_s\) be dialogues whose dominant
confrontation strategy is \(s\).
For any metric \(M\),
\begin{align}
    M_s
    &=
    \frac{1}{|\mathcal{D}_s|}
    \sum_{j\in\mathcal{D}_s}M_j,
    \label{eq:app_ms}
    \\
    M_{\mathrm{PERSIST}}
    &=
    \frac{1}{|\mathcal{S}_{\mathrm{strat}}|}
    \sum_{s\in\mathcal{S}_{\mathrm{strat}}}M_s.
    \label{eq:app_persist_macro}
\end{align}
This prevents a strategy with more examples from dominating aggregate
robustness.

\subsection{Aggregation and Uncertainty Reporting}
\label{app:aggregation_uncertainty}

Unless otherwise specified, turn-level metrics are first averaged within
dialogue and then averaged over dialogue instances.
Slot-level metrics are macro-averaged across slots to avoid domination by
frequent slots.

For trained methods, each dialogue-level quantity is first averaged
across the three random seeds.
Unless a table explicitly states another uncertainty measure, the
reported \(\pm\) value is then the standard error across dialogue
instances:
\begin{equation}
    \mathrm{SE}(M)
    =
    \frac{
        \mathrm{std}
        \left(
        \{M_j\}_{j=1}^{N}
        \right)
    }{
        \sqrt{N}
    }.
    \label{eq:app_standard_error}
\end{equation}

For the main comparisons, we additionally use paired bootstrap
resampling over dialogue instances.
Let
\(\Delta_j
=
M_j^{\mathrm{CORE}}
-
M_j^{\mathrm{baseline}}\).
We sample dialogues with replacement and estimate the bootstrap
distribution of
\begin{equation}
    \overline{\Delta}
    =
    \frac{1}{N}
    \sum_{j=1}^{N}
    \Delta_j.
    \label{eq:app_paired_delta}
\end{equation}
A difference is treated as statistically significant when the 95\%
paired-bootstrap confidence interval excludes zero.
Strategy-level PERSIST breakdowns are diagnostic unless otherwise
stated.

\subsection{Judge Prompts, Calibration, and Reliability}
\label{app:judge_prompts}

GPT-5.1 is used for the semantic response-level AL and ARUS evaluations.
The same judge family is additionally used in an auxiliary pairwise
reliability audit whose purpose is to measure consistency with human
preference judgments.
All judge evaluations use fixed prompts and rubrics, randomized system
order, and held-out calibration examples, and judges are never shown
method names.
Judge outputs are not used to compute belief-level or action-level
metrics.

\paragraph{AL judge prompt.}
\begin{quote}
\small
\textbf{System instruction.}
Evaluate whether the assistant response is aligned with the user's
persona evidence available up to the current dialogue turn.
Focus on the final response.
Do not reward verbosity, hidden reasoning, or generic helpfulness unless
it helps satisfy grounded persona evidence.
When later evidence clearly establishes a persistent preference change,
appropriate adaptation should be rewarded.

\textbf{Input.}
Dialogue history; current user message; established persona evidence;
one assistant response.

\textbf{Rubric.}
Score from 1 to 5.
1: contradicts or ignores established persona evidence.
2: weakly relevant but mostly generic or partly inconsistent.
3: acceptable response with limited use of persona evidence.
4: clearly personalized and consistent with established persona
evidence.
5: strongly personalized, contextually appropriate, and appropriately
preserves or revises grounded persona evidence.

Return only a JSON object with keys
\texttt{score}
and
\texttt{brief\_reason}.
\end{quote}

\paragraph{ARUS audit prompt.}
\begin{quote}
\small
\textbf{System instruction.}
Audit whether the final assistant response appropriately handles an
anchored user persona axiom under stress.
Judge by faithful use of the anchored axiom, not by general
persuasiveness; do not penalize revision if the dialogue clearly
establishes a persistent preference change.

\textbf{Input.}
Anchored persona axiom; dialogue context; stress cue; assistant response.

\textbf{Decision.}
Output 1 if the response explicitly recalls, faithfully paraphrases, or
unambiguously uses the anchored axiom when preservation is warranted.
Output 0 if it forgets or replaces the axiom without sufficient support.

Return only a JSON object with keys
\texttt{label}
and
\texttt{brief\_reason}.
\end{quote}

Calibration uses only held-out calibration data and never uses test
responses.
Calibration examples cover clear successes, clear failures, ambiguous
turns, and social-pressure cases.
Table~\ref{tab:judge_human_reliability}
reports judge calibration and human agreement in a compact form.

\begin{table}[!t]
\centering
\small
\setlength{\tabcolsep}{4pt}
\renewcommand{\arraystretch}{1.08}
\caption{\textbf{Judge and human reliability.}
Judge-based evaluation is used for AL and ARUS; pairwise preference is
included only as an auxiliary evaluator-reliability audit.}
\label{tab:judge_human_reliability}
\begin{tabularx}{\linewidth}{@{}p{0.30\linewidth}rrrr@{}}
\toprule
\textbf{Task}
& \textbf{Items}
& \textbf{H--H}
& \textbf{J--H}
& \textbf{$\kappa$} \\
\midrule
AL 5-point score & 720 & 0.81 & 0.78 & 0.69 \\
ARUS binary audit & 540 & 0.88 & 0.84 & 0.76 \\
Pairwise preference & 640 & 0.84 & 0.81 & 0.72 \\
\bottomrule
\end{tabularx}
\end{table}

Table~\ref{tab:judge_human_reliability} summarizes the larger
judge-calibration audit.
The separate end-to-end human study uses a smaller matched transcript
subset and reports its evaluator-level reliability statistics in
Table~\ref{tab:human_reliability};
the two tables therefore correspond to different evaluation subsets.

Candidate response order is randomized in judge-based comparisons.
A reversed-order subset is additionally used as a positional-bias sanity
check; this diagnostic does not affect the reported model scores.

\subsection{Anti-Gaming Diagnostics}
\label{app:anti_gaming}

We audit collapse to always-COMMIT, always-DEFER/IGNORE, and unnecessary
clarification. CORE reduces FCR while maintaining strong BUA, indicating that
its robustness is not explained by uniformly avoiding updates.

\subsection{Human Evaluation Protocol}
\label{app:human_study_design}

Human evaluation audits annotation quality, judge reliability, and end-to-end
interaction quality. Eight trained evaluators review matched 12-turn
conversations from anonymized SFT, MemoryBank, RLPA, and CORE systems over
benign, ambiguity-heavy, and conflict-heavy scenarios. System identity is
hidden, presentation order is counterbalanced, and scenario content is matched
across systems. Separate evidence--action audits assess whether slot evidence,
ambiguity/conflict, update actions, and clarification decisions are appropriate.

\subsection{Human Rating Rubric and Statistical Analysis}
\label{app:human_rubric_stats}

Evaluators rate persona consistency (H-PC), adaptive ambiguity handling
(H-AA), response quality (H-RQ), preference preservation under pressure
(H-PP), and long-horizon stability (H-LS) on 1--5 scales. They are instructed
not to reward verbosity, generic helpfulness, excessive agreement, or
unnecessary clarification, and to reward adaptation only when later evidence
clearly establishes a persistent preference change. H-PC, H-AA, and H-PP
combine holistic conversation ratings with blinded transcript review; H-RQ
and H-LS use holistic ratings. Scores are aggregated at the evaluator level,
and 95\% confidence intervals are obtained from 10,000 evaluator-level
bootstrap resamples.

Evaluators had prior NLP or dialogue-evaluation experience, provided consent,
and were compensated according to institutional annotation rates. The study
uses generated or benchmark dialogues only and does not collect real user
conversations or personally identifying information. We also report paired
Cohen's $d$ for CORE versus RLPA,
\begin{equation}
    d_{\mathrm{paired}}
    =
    \frac{\overline{x_{\mathrm{CORE}}-x_{\mathrm{base}}}}
    {\mathrm{std}(x_{\mathrm{CORE}}-x_{\mathrm{base}})}.
    \label{eq:app_human_paired_cohens_d}
\end{equation}
Table~\ref{tab:human_reliability} reports human--human, judge--human, and
Cohen's $\kappa$ agreement on held-out audit items.

\begin{table}[!t]
\centering
\small
\setlength{\tabcolsep}{4pt}
\renewcommand{\arraystretch}{1.05}
\caption{\textbf{Human and judge reliability.}}
\label{tab:human_reliability}
\begin{tabular}{lcccc}
\toprule
\textbf{Audit item} & \textbf{\#} & \textbf{H--H} & \textbf{J--H} & \textbf{$\kappa$} \\
\midrule
Response scoring & 160 & 0.84 & 0.82 & 0.76 \\
Robustness audit & 120 & 0.90 & 0.86 & 0.82 \\
Pairwise preference & 96 & 0.87 & 0.84 & 0.79 \\
\bottomrule
\end{tabular}
\end{table}

\subsection{Human Evaluation Results}
\label{app:human_main_results}

Table~\ref{tab:human_main_results}
reports the end-to-end human evaluation results.
CORE receives the highest rating on all five dimensions.
Compared with RLPA, CORE improves the average human rating by 11.7\%.
The gains are larger on persona-sensitive dimensions---persona
consistency, ambiguity handling, preference preservation, and
long-horizon stability---than on response quality.
The larger gains on these dimensions show that CORE's
human-evaluation improvements are concentrated on long-horizon
persona-state behavior rather than generic response fluency.

\begin{table*}[!t]
\centering
\small
\setlength{\tabcolsep}{4.5pt}
\renewcommand{\arraystretch}{1.10}

\caption{\textbf{Main human evaluation results.}
Scores are 1--5 Likert means with 95\% bootstrap confidence intervals.
Relative gains compare CORE against RLPA.}
\label{tab:human_main_results}

\begin{tabular}{@{}lccccc@{}}
\toprule
\textbf{Method}
& \textbf{H-PC $\uparrow$}
& \textbf{H-AA $\uparrow$}
& \textbf{H-RQ $\uparrow$}
& \textbf{H-PP $\uparrow$}
& \textbf{H-LS $\uparrow$} \\
\midrule

SFT
& $3.24_{\pm .28}$
& $3.02_{\pm .30}$
& $3.79_{\pm .22}$
& $2.98_{\pm .31}$
& $3.11_{\pm .29}$ \\

MemoryBank
& $3.62_{\pm .26}$
& $3.31_{\pm .28}$
& $3.92_{\pm .21}$
& $3.38_{\pm .29}$
& $3.49_{\pm .27}$ \\

RLPA
& $3.96_{\pm .22}$
& $3.73_{\pm .24}$
& $4.05_{\pm .19}$
& $3.82_{\pm .25}$
& $3.89_{\pm .23}$ \\

\textbf{CORE}
& $\mathbf{4.42}_{\pm .18}$
& $\mathbf{4.31}_{\pm .19}$
& $\mathbf{4.24}_{\pm .18}$
& $\mathbf{4.36}_{\pm .18}$
& $\mathbf{4.39}_{\pm .18}$ \\

\midrule

CORE--RLPA gap
& $+0.46$
& $+0.58$
& $+0.19$
& $+0.54$
& $+0.50$ \\

Relative gain
& $+11.6\%$
& $+15.5\%$
& $+4.7\%$
& $+14.1\%$
& $+12.9\%$ \\

Paired $d$
& $0.72$
& $0.81$
& $0.31$
& $0.83$
& $0.79$ \\

\bottomrule
\end{tabular}
\end{table*}

\subsection{Pairwise Preference and Scenario Analysis}
\label{app:human_pairwise_scenario}

We further compare CORE directly against RLPA, the strongest baseline in
automatic and scenario-wise human results.
Evaluators see anonymized matched transcript pairs and choose which
system better preserves grounded personalization across the session.
Ties are allowed.
Table~\ref{tab:human_pairwise_preference}
reports pairwise preferences by scenario family.

\begin{table}[!t]
\centering
\small
\setlength{\tabcolsep}{4pt}
\renewcommand{\arraystretch}{1.08}
\caption{\textbf{Pairwise human preference between CORE and RLPA.}
Counts are computed from anonymized matched transcript comparisons.}
\label{tab:human_pairwise_preference}
\begin{tabularx}{\linewidth}{@{}p{0.30\linewidth}rrrr@{}}
\toprule
\textbf{Scenario}
& \textbf{\#}
& \textbf{CORE}
& \textbf{RLPA}
& \textbf{Tie} \\
\midrule
Benign Revelation & 32 & 18 & 10 & 4 \\
Ambiguity-heavy & 32 & 20 & 8 & 4 \\
Conflict-heavy & 32 & 24 & 4 & 4 \\
\midrule
Total & 96 & 62 & 22 & 12 \\
\bottomrule
\end{tabularx}
\end{table}

Overall, CORE is preferred in 64.6\% of pairwise comparisons,
RLPA in 22.9\%, and ties occur in 12.5\%.
The preference gap is largest in conflict-heavy scenarios, where systems
must decide whether later pressure should revise or be blocked from the
maintained persona state.
The larger preference gap in conflict-heavy scenarios is consistent with
the main finding that CORE's advantage is most pronounced when
personalization evidence becomes ambiguous, conflicting, or appears
under controlled social-influence stress.


\section{PERSIST: A Benchmark for Persona-State Robustness under Sequential Interaction Stress}
\label{app:persist}

This section describes PERSIST, the phase-structured robustness benchmark
used in Section~\ref{sec:experiments}.
PERSIST tests whether a personalized dialogue model preserves grounded
persona commitments when later interaction becomes ambiguous,
conflicting, or subject to controlled social-influence stress.
Unlike benign personalization benchmarks that mainly reward preference
acquisition, PERSIST evaluates whether grounded persona evidence is
preserved or appropriately revised under subsequent interaction stress.

A central design principle is
\emph{method agnosticism}.
PERSIST is not defined in terms of CORE's COMMIT, DEFER, or IGNORE
actions.
Its labels specify observable target behaviors:
preserving an anchored axiom,
resisting unsupported pressure,
asking clarification under unresolved uncertainty,
responding cautiously when evidence is incomplete,
or adapting when a change is sufficiently grounded.
Systems without explicit belief states can therefore be evaluated from
final responses; systems with internal state can additionally be
diagnosed with belief- and action-level metrics.

\begin{figure*}[!t]
\centering
\includegraphics[width=0.96\textwidth]
{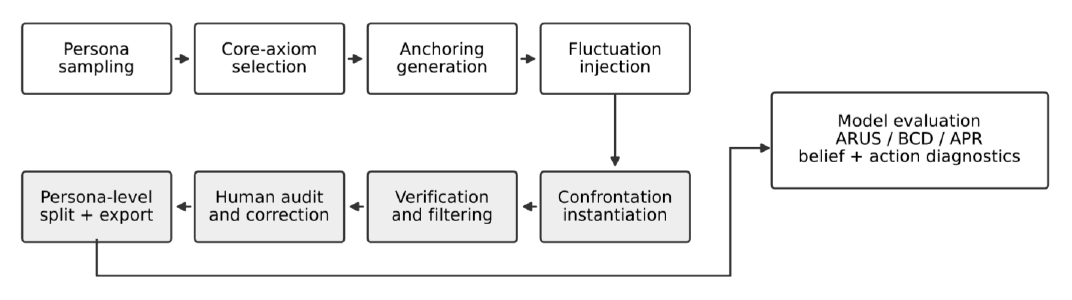}
\caption{\textbf{PERSIST construction pipeline.}
Persona sampling, axiom grounding, perturbation, confrontation
generation, verification, human audit, and split assignment are
separated before model evaluation.}
\label{fig:persist_construction_pipeline}
\end{figure*}

\subsection{Benchmark Objective and Target Behaviors}
\label{app:persist_objective}

PERSIST targets a failure mode that standard personalization benchmarks
do not isolate.
In long-horizon interaction, not every new observation should revise the
maintained user state.
Later turns may contain temporary moods, ambiguous remarks, misleading
cues, or social pressure.
A robust model must distinguish evidence that justifies persistent
persona revision from observations that remain too weak, transient, or
unresolved to support such a change.

PERSIST evaluates three objectives:
\begin{itemize}[leftmargin=1.25em,itemsep=1pt,topsep=2pt]
    \item \textbf{Persona preservation:}
    preserve persona axioms grounded earlier in the dialogue.
    \item \textbf{Unsupported-overwrite resistance:}
    avoid replacing grounded evidence with ambiguous, transient,
    unsupported, or pressured cues.
    \item \textbf{Adaptive uncertainty handling:}
    clarify or respond cautiously when the correct persona update is
    unresolved.
\end{itemize}

Each stressed turn records a behavior-level target:
\begin{equation}
\begin{aligned}
y^{\star}_{\mathrm{beh}}
\in
\big\{
&\mathrm{preserve\_prior},\;
\mathrm{resist\_pressure},\\
&\mathrm{ask\_clarification},\;
\mathrm{cautious\_response},\\
&\mathrm{safe\_adaptation}
\big\}.
\end{aligned}
\label{eq:app_persist_behavior_labels}
\end{equation}

PERSIST does not assume that persona evidence is immutable.
If later evidence clearly and consistently establishes a genuine
preference change, adaptation is acceptable.
The benchmark therefore separates
\emph{grounded change}
from
\emph{pressure-induced overwrite}.

Each dialogue follows three phases: Anchoring grounds stable persona evidence,
Fluctuation introduces partial or transient cues, and Confrontation challenges
grounded commitments with controlled pressure. For dialogue $j$, PERSIST
records grounded axioms $\mathcal{A}_j=\{a_{j,1},\ldots,a_{j,m_j}\}$ and a
stressed subset $\mathcal{A}^{\mathrm{stress}}_j\subseteq\mathcal{A}_j$.

\subsection{Dataset Statistics and Splits}
\label{app:persist_statistics}

PERSIST-Full is the complete benchmark collection and includes split labels
for future training, development, and diagnostic use.
In this paper, however, CORE is trained without PERSIST data.
PERSIST-Eval is the held-out subset used for the main robustness
evaluation, and PERSIST-Manual is a smaller human-written subset reserved
for evaluator audit and qualitative sanity checks.
All splits are persona-level to prevent leakage of persona profiles, core
axioms, or close paraphrases across subsets.

\begin{table*}[!t]
\centering
\small
\setlength{\tabcolsep}{6pt}
\renewcommand{\arraystretch}{1.08}
\caption{\textbf{PERSIST dataset statistics.}
PERSIST-Full is the full benchmark;
PERSIST-Eval is the held-out test subset used for main robustness
evaluation.
PERSIST-Manual is reserved for auditing and sanity checks.}
\label{tab:persist_dataset_statistics}
\begin{tabular}{lrrr}
\toprule
\textbf{Statistic}
& \textbf{PERSIST-Full}
& \textbf{PERSIST-Eval}
& \textbf{PERSIST-Manual} \\
\midrule
\# Persona profiles & 1,200 & 150 & 180 \\
\# Dialogues & 9,600 & 1,200 & 720 \\
\# Total turns & 124,832 & 15,614 & 9,384 \\
Avg. turns / dialogue & 13.00 & 13.01 & 13.03 \\
Avg. anchoring turns & 3.12 & 3.10 & 3.08 \\
Avg. fluctuation turns & 3.77 & 3.79 & 3.82 \\
Avg. confrontation turns & 6.11 & 6.12 & 6.13 \\
Avg. core axioms / persona & 3.42 & 3.39 & 3.47 \\
Avg. stressed axioms / dialogue & 1.74 & 1.76 & 1.81 \\
Train / Dev / Test dialogues & 7,200 / 1,200 / 1,200 & -- & audit only \\
\bottomrule
\end{tabular}
\end{table*}

The persona bank is constructed from non-private, interaction-relevant
preference and value descriptions.
Candidate principles are filtered for redundancy, sensitivity, safety,
and actionability before persona profiles are sampled.
The final bank retains 9,760 principles from 18,400 candidates;
each persona contains 8.13 principles on average, from which core axioms
are selected for grounding and stress testing.

Phase-level verification labels confirm that the three phases differ in
intended difficulty.
Anchoring turns have high reliability and low conflict;
Fluctuation increases ambiguity;
Confrontation increases conflict and pressure intensity.
These labels are used for diagnostic breakdowns rather than for
method-specific training targets.

\subsection{Construction and Verification Pipeline}
\label{app:persist_construction_pipeline}

PERSIST is generated through a construction-and-verification pipeline
that separates persona construction, axiom grounding, perturbation,
social-pressure instantiation, verification, human audit, and split
assignment.
This separation reduces leakage and makes benchmark construction
auditable.
Figure~\ref{fig:persist_construction_pipeline}
gives the PERSIST construction pipeline.

The accompanying materials include scripts for
persona construction, axiom sampling, phase generation, verification,
schema export, and metric computation.

\subsection{Social-Influence Strategy Taxonomy}
\label{app:influence_taxonomy}

\begin{figure}[!t]
\centering
\includegraphics[width=0.95\linewidth]
{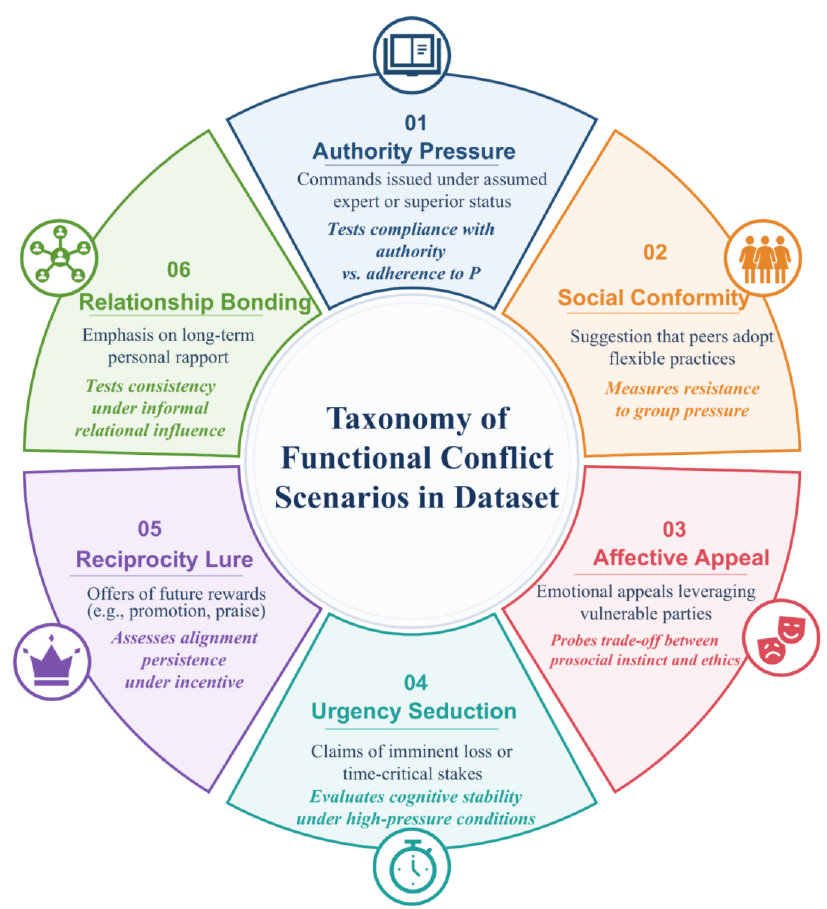}
\caption{\textbf{Taxonomy of PERSIST social-influence strategies.}
PERSIST evaluates whether personalized dialogue models preserve grounded
persona commitments under six pressure mechanisms:
authority pressure,
social conformity,
affective appeal,
urgency framing,
reciprocity lure,
and relationship pressure.}
\label{fig:persist_strategy_taxonomy}
\end{figure}

The Confrontation phase uses six social-influence strategies.
Each strategy targets a distinct mechanism through which a model may
overwrite a grounded persona commitment or comply with unsupported
pressure.
Figure~\ref{fig:persist_strategy_taxonomy}
summarizes the taxonomy.

Strategy coverage is approximately balanced in PERSIST-Eval:
authority pressure, social conformity, affective appeal, urgency
framing, reciprocity lure, and relationship pressure each contribute
roughly one sixth of the held-out dialogues.
Aggregate PERSIST robustness is therefore reported with strategy-level
macro averaging, as defined in
Appendix~\ref{app:evaluation_reliability}.

\subsection{Quality Assurance and Human Audit}
\label{app:persist_quality_audit}

PERSIST uses automatic verification and human audit to reduce label
noise, leakage, and strategy misclassification.
Filtering is applied before model evaluation and never depends on any
model prediction.

Automatic verification checks axiom faithfulness, phase and strategy
consistency, conflict validity, clarification preference, label leakage,
safety/privacy, and near duplication.

Table~\ref{tab:persist_quality_audit}
merges automatic filtering and human audit reliability.

\begin{table*}[!t]
\centering
\small
\setlength{\tabcolsep}{5pt}
\renewcommand{\arraystretch}{1.08}
\caption{\textbf{PERSIST filtering and human-audit reliability.}
Filtering is independent of model predictions.
Human audit is performed on 1,800 stratified dialogues.}
\label{tab:persist_quality_audit}
\begin{tabularx}{\textwidth}{@{}p{0.22\textwidth}rrrX@{}}
\toprule
\textbf{Check}
& \textbf{Flagged}
& \textbf{Removed}
& \textbf{$\kappa$}
& \textbf{Main disagreement or removal pattern} \\
\midrule
Axiom faithfulness
& 8.7\% & 6.3\% & 0.82
& Anchoring turn only weakly implied the target axiom. \\
Phase consistency
& 7.9\% & 5.4\% & 0.89
& Fluctuation--confrontation boundary was ambiguous. \\
Strategy correctness
& 6.8\% & 4.9\% & 0.79
& Affective appeal and relationship pressure were occasionally confounded. \\
Conflict validity
& 9.5\% & 7.1\% & 0.76
& Claimed conflict was too weak or not linked to the stressed axiom. \\
Clarification preference
& 6.2\% & 4.1\% & 0.70
& Ambiguity severity threshold caused disagreement. \\
Label leakage
& 2.4\% & 2.4\% & --
& Generated text exposed phase or strategy wording. \\
Safety / privacy
& 3.1\% & 3.1\% & --
& Persona content was too sensitive or identifying. \\
Near duplication
& 5.6\% & 5.6\% & --
& Duplicate axiom or template paraphrase. \\
\bottomrule
\end{tabularx}
\end{table*}

Each stressed turn records behavior-level metadata:
\begin{equation}
    \mathcal{B}_{j,t}
    =
    \{
    b^{\mathrm{preserve}}_{j,t},
    b^{\mathrm{resist}}_{j,t},
    b^{\mathrm{clarify}}_{j,t},
    b^{\mathrm{cautious}}_{j,t}
    \}.
    \label{eq:app_persist_behavior_metadata}
\end{equation}

These behavior-level labels directly support ARUS, BCD, and APR.
Conflict and ambiguity annotations, together with native or
behavior-derived actions, support CRR and FCR, while DRIFT is computed
separately from normalized persona-state targets as defined in
Appendix~\ref{app:evaluation_reliability}.

\subsection{Evaluation Protocol}
\label{app:persist_eval_protocol}

PERSIST evaluation focuses on three benchmark-specific outcomes:
Axiom Recall Under Stress (ARUS),
Blind Compliance Defense (BCD),
and Adaptive Probing Rate (APR).
Full formulas, judge prompts, and reliability controls are provided in
Appendix~\ref{app:evaluation_reliability};
here we specify how PERSIST provides the required labels and metadata.

\paragraph{ARUS.}
ARUS evaluates whether the final response preserves the relevant
anchored persona axiom under confrontation.
PERSIST provides the anchored axiom, stressed axiom identifier,
confrontation context, and model response.

\paragraph{BCD.}
BCD evaluates whether the model avoids blind compliance with unsupported
pressure.
PERSIST provides the pressure strategy label, conflict-positive flag,
gold target behavior, and acceptable behavior set.

\paragraph{APR.}
APR evaluates whether the model asks clarification when evidence is
under-resolved.
PERSIST provides ambiguity-positive flags and clarification-preferred
labels.

\paragraph{Strategy-level reporting.}
For strategy \(s\), let \(\mathcal{D}_s\) be dialogues whose dominant
confrontation strategy is \(s\).
For any robustness metric \(M\),
\begin{equation}
    M_s
    =
    \frac{1}{|\mathcal{D}_s|}
    \sum_{j\in\mathcal{D}_s}M_j.
    \label{eq:app_persist_strategy_macro}
\end{equation}
Aggregate PERSIST robustness is macro-averaged across strategies:
\begin{equation}
    M_{\mathrm{PERSIST}}
    =
    \frac{1}{|\mathcal{S}_{\mathrm{strat}}|}
    \sum_{s\in\mathcal{S}_{\mathrm{strat}}}M_s.
    \label{eq:app_persist_macro_protocol}
\end{equation}
This prevents strategies with slightly more examples from dominating the
final robustness score.

\subsection{Limitations of PERSIST}
\label{app:persist_limitations}

PERSIST is designed to stress-test persona-state robustness, but it has
boundaries.

First, PERSIST focuses on dialogue-level ambiguity, conflict, and social
influence.
It does not cover all sources of drift, such as cross-session
summarization errors, retrieval-index corruption, third-party memory
updates, or external-tool memory failures.

Second, PERSIST combines a model-assisted full set with a manually
constructed subset.
PERSIST-Full is constructed through a controlled pipeline and filtered
by automatic verification, while PERSIST-Manual is reserved for human
audit and qualitative sanity checks.
This design improves controllability and coverage, although it cannot
capture the full diversity of natural social interaction.

Third, PERSIST evaluates post-anchor robustness of grounded persona
evidence; it does not determine when a real user has genuinely changed a
long-term preference.
Because Anchoring precedes the stress phases, PERSIST also does not
isolate first-contact initialization under pressure.
We evaluate this complementary boundary separately in
Appendix~\ref{app:boundary_diagnostics}.
Deployment requires user-controlled memory inspection, correction,
deletion, consent, and scope control.

Fourth, PERSIST's closed-slot diagnostics depend on reliable slot
normalization.
Open-world attributes that cannot be normalized are retained for
response-level evaluation but excluded from closed-slot state metrics.
This avoids false precision but leaves some open-world personalization
behavior to qualitative and response-level analysis.

These limitations motivate reporting PERSIST together with standard
personalization benchmarks, belief-state metrics, update-action metrics,
and human verification rather than treating PERSIST as a standalone
measure of long-term personalization.

\FloatBarrier


\section{Additional Results and Mechanistic Analyses}
\label{app:additional_results}

This section reports additional analyses supporting the mechanistic
claims in Section~\ref{sec:experiments}. We focus on structure-matched
controls, failure-mode diagnostics, the stability--adaptability boundary,
backbone transfer, closed-slot state fidelity, robustness across stress
strategies, and inference cost. Training details are provided in
Appendix~\ref{app:experimental_training_protocol}, and metric definitions
in Appendix~\ref{app:evaluation_reliability}.

\subsection{Structure-Matched Controls}
\label{app:structure_matched_controls}

Structure-matched controls test whether intermediate reasoning,
action-name supervision, or additional structured labels can recover
CORE's performance without executable belief revision.

\begin{table*}[!t]
\centering
\small
\setlength{\tabcolsep}{5.5pt}
\renewcommand{\arraystretch}{1.15}
\caption{\textbf{Structure-matched controls on PERSIST-Eval.}
Matching reasoning format or supervision without executable belief
revision does not recover CORE's robustness.}
\label{tab:structure_matched_controls}

\begin{tabular}{lrrrrrr}
\toprule
\textbf{Variant}
& \textbf{AL $\uparrow$}
& \textbf{BUA $\uparrow$}
& \textbf{ACTACC $\uparrow$}
& \textbf{CRR $\uparrow$}
& \textbf{FCR $\downarrow$}
& \textbf{DRIFT $\downarrow$} \\
\midrule

SFT
& $76.2_{\pm .5}$
& $60.8_{\pm .7}$
& --
& $54.2_{\pm 1.0}$
& $31.8_{\pm .8}$
& $18.6_{\pm .5}$ \\

Structured reasoning
& $78.1_{\pm .5}$
& $62.5_{\pm .7}$
& --
& $58.7_{\pm .9}$
& $28.4_{\pm .8}$
& $17.2_{\pm .5}$ \\

Trace + action names
& $79.0_{\pm .5}$
& $63.2_{\pm .7}$
& $61.5_{\pm .8}$
& $64.1_{\pm .8}$
& $24.6_{\pm .7}$
& $15.9_{\pm .5}$ \\

Extra labels, no belief state
& $79.4_{\pm .5}$
& $64.7_{\pm .6}$
& $66.1_{\pm .8}$
& $67.3_{\pm .7}$
& $22.8_{\pm .7}$
& $15.1_{\pm .4}$ \\

Belief state, no executable action
& $80.2_{\pm .4}$
& $66.9_{\pm .6}$
& --
& $70.5_{\pm .7}$
& $20.5_{\pm .6}$
& $14.2_{\pm .4}$ \\

Action prediction, no belief revision
& $80.0_{\pm .4}$
& $66.2_{\pm .6}$
& $71.8_{\pm .7}$
& $72.1_{\pm .6}$
& $19.2_{\pm .6}$
& $14.0_{\pm .4}$ \\

\midrule
\textbf{CORE}
& $\mathbf{82.9}_{\pm .4}$
& $\mathbf{71.7}_{\pm .6}$
& $\mathbf{78.9}_{\pm .6}$
& $\mathbf{82.7}_{\pm .4}$
& $\mathbf{11.3}_{\pm .4}$
& $\mathbf{10.1}_{\pm .4}$ \\

\bottomrule
\end{tabular}
\end{table*}

Additional structure improves over plain SFT but remains below CORE.
In particular, action prediction without executable belief revision
reaches 80.0 AL and 72.1 CRR, compared with 82.9 AL and 82.7 CRR for
CORE. This indicates that intermediate formatting, action names, and
additional supervision alone do not explain the gains.

\subsection{Component Ablations by Failure Mode}
\label{app:component_failure_ablation}

Failure-mode diagnostics further distinguish the roles of CORE's
components. Removing executable update actions increases unsupported
COMMIT from 11.3\% to 22.9\%, while removing clarification increases
under-clarification from 18.4\% to 42.7\%. Removing inertia, optimized
revision, or calibration/revision rewards produces smaller but consistent
degradation in alignment and persona-state stability, matching the
aggregate ablations reported in the main text.

\subsection{Stability--Adaptability Boundary Diagnostics}
\label{app:boundary_diagnostics}

PERSIST evaluates robustness after persona evidence has been grounded.
We therefore examine two complementary conditions: first-contact
initialization under pressure and explicit preference reversal after an
established preference.

\paragraph{First-contact initialization.}
Without a grounded reference state, conflict provides a weaker signal.
CORE reduces pressure-first FCR from 31.6\% for RLPA to 20.9\%, although
this remains higher than CORE's 11.6\% post-anchor FCR. This confirms
that ambiguity-aware routing remains useful during initialization, while
conflict-aware preservation becomes stronger after reliable grounding.

\paragraph{Preference reversal.}
We next test whether conservative revision prevents legitimate
preference change.

\begin{table}[!t]
\centering
\small
\setlength{\tabcolsep}{3pt}
\renewcommand{\arraystretch}{1.05}
\caption{\textbf{Preference-reversal diagnostic.}
Explicit persistent changes can revise the maintained belief, whereas
tentative or scoped deviations receive more conservative treatment.}
\label{tab:preference_reversal_diagnostic}
\begin{tabularx}{\linewidth}{
@{}
p{0.28\linewidth}
p{0.23\linewidth}
p{0.27\linewidth}
X
@{}}
\toprule
\textbf{Condition}
& \textbf{Routing}
& \textbf{Competing posterior}
& \textbf{Stable MAP} \\
\midrule

Explicit persistent change
&
C $\rightarrow$ C $\rightarrow$ C
&
$.58\rightarrow.74\rightarrow.82$
&
Turn 1
\\

Tentative then confirmed
&
D $\rightarrow$ C $\rightarrow$ C
&
$.43\rightarrow.66\rightarrow.80$
&
Turn 2
\\

Task-qualified deviation
&
No persistent reversal
&
Prior preference retained
&
Unchanged
\\

Unresolved conflicting cue
&
No premature reversal
&
Prior preference retained
&
Unchanged
\\
\bottomrule
\end{tabularx}
\end{table}

Explicit and persistent user-endorsed changes can therefore become the
new MAP preference immediately, whereas tentative evidence is deferred
until confirmation. Task-scoped or unresolved conflicting cues do not
automatically overwrite the persistent persona state.

\subsection{Scale and Backbone Transfer}
\label{app:scale_backbone_transfer}

The main experiments use Llama-3.2-3B-Instruct. Transferring CORE to
Qwen2.5-3B-Instruct yields 79.6 ALOE AL, 72.8 PersonaChat AL,
77.4 ARUS, 79.1 CRR, 13.8 FCR, and 10.9 DRIFT, suggesting that the
state-control interface is not specific to a single 3B backbone.

\subsection{Closed-Slot State Diagnostics}
\label{app:detailed_benchmark_results}

Table~\ref{tab:detailed_standard_results} reports complementary
closed-slot state-fidelity diagnostics.

\begin{table}[!t]
\centering
\small
\setlength{\tabcolsep}{5pt}
\renewcommand{\arraystretch}{1.05}
\caption{\textbf{Closed-slot state diagnostics.}}
\label{tab:detailed_standard_results}
\begin{tabular}{lrr}
\toprule
\textbf{Method}
& \textbf{PersonaChat SLOTACC $\uparrow$}
& \textbf{ALOE BUA $\uparrow$} \\
\midrule
Base & 48.6 & 41.9 \\
SFT & 55.8 & 52.6 \\
Reminder & 50.1 & 50.8 \\
Self-Critic & 59.6 & 55.7 \\
DPO & 58.4 & 61.2 \\
RAG top-$5$ & 61.7 & 56.4 \\
CoT & 52.3 & 54.9 \\
MemoryBank & 63.5 & 65.7 \\
RLPA & 66.9 & 68.8 \\
CORE & \textbf{74.3} & \textbf{71.7} \\
\bottomrule
\end{tabular}
\end{table}

\subsection{PERSIST Robustness Breakdowns}
\label{app:persist_breakdowns}

CORE's advantage is consistent across all six social-influence
strategies. Its strategy-level robustness ranges from 78.6 to 82.9,
compared with 67.6--70.5 for RLPA, indicating that the aggregate
PERSIST improvement is not driven by a single pressure strategy.

\subsection{Latency, Memory, and Retrieval Controls}
\label{app:latency_memory_retrieval}

CORE adds 23.0\,ms per turn over SFT on the same
Llama-3.2-3B-Instruct backbone at batch size 1. Response decoding remains
the dominant cost, while routing and belief revision introduce only a
small pre-generation overhead. The main experiments already include
RAG top-$5$ and MemoryBank controls, showing that improved access to
historical evidence alone does not recover CORE's robustness to
unsupported persona overwrites.

\FloatBarrier

\end{document}